\UseRawInputEncoding
\documentclass[journal=jacsat,manuscript=article]{achemso}

\usepackage[version=3]{mhchem} 
\usepackage{multirow}
\usepackage{makecell}
\usepackage{cellspace}
\usepackage{booktabs}
\usepackage{amsmath} 
\usepackage{amssymb} 
\usepackage{enumitem}
\usepackage{algorithm}
\usepackage{hyperref}
\usepackage{url}
\usepackage{adjustbox}
\usepackage{algpseudocode}
\usepackage{lipsum}\usepackage{tabularx}
\usepackage{booktabs}
\usepackage{xcolor}

\author{Vani Nigam}
\affiliation{Department of Materials Science and Engineering, Carnegie Mellon University, 5000 Forbes Avenue, Pittsburgh, PA 15213, USA}
\author{Achuth Chandrasekhar}
\affiliation{Department of Mechanical Engineering, Carnegie Mellon University, 5000 Forbes Avenue, Pittsburgh, PA 15213, USA}
\author{Amir Barati Farimani}
\affiliation{Department of Mechanical Engineering, Carnegie Mellon University, 5000 Forbes Avenue, Pittsburgh, PA 15213, USA}
\email{barati@cmu.edu}

\title[An \textsf{achemso} demo]
  {Polymer-Agent: Large Language Model Agent for Polymer Design}

\abbreviations{IR,NMR,UV}
\keywords{American Chemical Society, \LaTeX}

\mciteErrorOnUnknownfalse
\begin{document}







\begin{abstract}
On-demand Polymer discovery is essential for various industries, ranging from biomedical to reinforcement materials. Experiments with polymers have a long trial-and-error process leading to use of extensive resources. For these processes, machine learning has accelerated scientific discovery at the property prediction and latent space search fronts. However, laboratory researchers cannot readily access codes and these models to extract individual structures and properties due to infrastructure limitations. We present a closed-loop polymer structure-property predictor integrated in a terminal for early-stage polymer discovery. The framework is powered by LLM reasoning to provide users with property prediction, property-guided polymer structure generation, and structure modification capabilities. The SMILES sequences are guided by the synthetic accessibility score and the synthetic complexity score (SC Score) to ensure that polymer generation is as close as possible to synthetically accessible monomer-level structures. This framework addresses the challenge of generating novel polymer structures for laboratory researchers, thereby providing computational insights into polymer research.
\end{abstract}

\section{Introduction}
Polymer materials present a wide range of tunable biodegradation, mechanical properties, porosity, and surface-to-volume ratio \cite{terzopoulou_biocompatible_2022}. These materials are used in applications to increase durability in solar cells through a cross-linking strategy in three-dimensional hyperbranched polymers\cite{li_boosting_2025}. Polymer scaffolds are utilized in cardiovascular operations, employing multiple polymer scaffolds to mimic the human myocardium and improve hydrophilicity and biodegradability\cite{toh_polymer_2021}. Conjugated polymers (CPs) are organic semiconductor materials with large pi-conjugated backbones that allow a wide light absorption range and light harvesting for antitumor therapy, differentiation, and killing of pathogenic bacteria, and intracellular imaging\cite{noauthor_sensing_nodate}. Polyethylene oxide (PEO)-based electrolytes are examples of polymer-based electrolytes for uniform, fast, and stable migration/diffusion behavior\cite{li_regulating_2024}. 
Primary and secondary structures of polymers play a pivotal role in the variety of the above characteristics of the wide-ranging soft materials including thermoplastic elastomers, thermosets, membranes, hydrogels, emulsions, etc. The field of polymer chemistry has seen tremendous growth in molecular precision and the synthesis of macromolecules with a range of controlled composition, chain ends, topology, molecular weight, and polymerization \cite{lutz_precision_2016}. Polymers have a vast parameter space, exhibiting variations in molecular chains, compositional polydispersity, sequence randomness, multi-level structures\cite{meijer_mechanical_2005}. 
Researchers work on fabricating materials by developing synthesis routes that target specific properties. Conjugated polymer nanomaterials are reactions of dimers resulting from radical cations by oxidizing monomers. The initiation of polymerization involves three routes: chemical, electrochemical, and photo-induced oxidation\cite{nguyen_recent_2016}. Similarly, the process for synthesizing conductive polymer gels (CPGs) involves cross-linking of molecules with multiple functional groups to form 3D molecular networks to improve the conductivity of these networks. These materials have applications in bioelectronics, energy storage, and conversion devices due to their conductive structural properties \cite{zhao_multifunctional_2017}.
However, this trial-and-error method for designing novel polymer structures is costly and laborious. This necessitates the exploration of more efficient methods to explore the parameters of polymer properties, thereby accelerating the research process. \cite{zhao_machine-learning-assisted_2024}
ML-based experimental recommendations have guided material discovery in phase change memory alloys, high entropy alloys, photovoltaics, and porous membrane design.\cite{adams_human---loop_2024}\cite{cao_machine_2024}. The integration of Large Language Models (LLMs) in polymer design involves training or fine-tuning existing models using materials databases that require multimodal learning, high-cost, and energy-intensive methods. \cite{badrinarayanan_multi-peptide_2025}\cite{liao_inverse_2025} \cite{zheng_ai-guided_2025}\cite{sahu_encoder-decoder_2025}\cite{choudhary_atomgpt_2024}\cite{balaji_gpt-molberta_2023} 
Multiple research studies, conducted in various capacities, have focused on data-driven material discovery. For example, Variational Autoencoders (VAE) are used to encode high-dimensional polymer structures into continuous, low-dimensional latent spaces, which enable high-throughput search. Graph neural networks are used to represent molecules and thereby learn their chemical bonds. Bayesian optimization has been experimented with, to traverse the latent space to target polymers with desired properties \cite{vogel_inverse_2024}. As a self-supervised model, MolCLR uses a large unlabeled dataset to train and learn molecular similarities. \cite{wang_molecular_2022}  Property prediction tasks have also gained momentum in Large Language Model research, which treats polymer SMILES as sequences to leverage progress in natural language processing tasks and uses regression heads to use labeled property databases.\cite{xu_transpolymer_2023}\cite{kuenneth_polybert_2023} 

Despite multiple steps in different directions, methods remain isolated and task-specific. These isolated implementations limit the scalability and efficiency of current pipelines. A polymer discovery experiment requires the seamless integration of various computational methods.\cite{pak_agentic_2025} Compared to using only computational methods for material simulations\cite{barati_farimani_fast_2024}, ML-based approaches can offer fast, high-throughput screening for polymer discovery, prediction, and design. Inverse design of materials inherits an extremely vast search space, which can be navigated using one of the following methods: (1) High-throughput virtual screening, (2) Global optimization, (3) Generative models. \cite{wang_inverse_2022} In research, the inverse design of specific polymers and the statistical design of experiments using explainable AI are implemented. \cite{dangayach_machine_2025} Additionally, there have been advancements in Large Language Models (LLMs), which are fine-tuned on a material database and utilize LoRA, as well as API from LLMs like OpenAI (\url{https://openai.com/api/}) and Gemini (\url{https://ai.google.dev/gemini-api/docs}) , to facilitate polymer-specific conversations with the LLM. PolySea, a domain-specific LLMs for polymer informatics, achieves a classification accuracy of 79\% in thermal stability prediction.\cite{qiu_introducing_2025} Fine-tuned GPT-3.5 can achieve predictive accuracies of solubility of up to 90\%. \cite{agarwal_polymer_2025} 
However, the use of stand-alone LLMs and neural networks for scientific discovery suffers from cognitive deficiency, i.e., the inability to decide whether or not to move to the next step of the procedure. Agentic AI, on the other hand, scales horizontally by employing an LLM as the brain of the process and external tools to guide and monitor every step the agent takes.\cite{tran_multi-agent_2025}\cite{chandrasekhar_automating_2025} Many interdisciplinary applications have used agentic AI to present users with contextual querying over domain knowledge, subjective tools for diverse tasks using foundational models and APIs etc. \cite{chandrasekhar_amgpt_2024}\cite{liu_mcpeval_2025}\cite{chandrasekhar_nanogpt_2025}\cite{noauthor_taskmatrixai_nodate}
Within the agentic workflows, the past research shows meaningful integrations with manually integrated APIs, hardcoded Pythonic algorithms, and frameworks like ReAct\cite{yao_react_2023}. However, these frameworks are brittle due to schema enforcement and API costs during tool invocation\cite{mastouri_making_2025}. 
In this work, we introduce Polymer-Agent, an LLM-powered agent framework designed to leverage agentic AI to keep a human in the loop during polymer design. The agent performs two essential tasks: predicting a polymer SMILES property and generating target-motivated polymer SMILES. The Model Context Protocol (MCP: \url{https://www.anthropic.com/news/model-context-protocol}) servers enable a simulation researcher to access the generated results in an integrated terminal using natural language queries.

\section{Methods}

\subsection{Polymer Informatics}
\subsubsection{Generative Models}
In recent years, applications of Transformers\cite{vaswani_attention_2017} have advanced significantly in natural language processing (NLP) and AI for science. Transformer-based models have emerged for molecule property predictions, processing reactions, and as a structure-agnostic model for text string representation in SMILES-BERT\cite{wang_smiles-bert_2019}, ChemBERTa\cite{chithrananda_chemberta_2020}, Molecular Transformer\cite{schwaller_molecular_2019}, Moformer\cite{cao_moformer_2023}, PolyRetro \cite{agarwal_polyretro_2025}, MOFGPT\cite{badrinarayanan_mofgpt_2025}.
TransPolymer\cite{xu_transpolymer_2023} is a transformer-based language model for polymer property predictions, reaching test R² of 0.92-0.93 in the polymer properties of bulk bandgap, 0.69 in electric conductivity, 0.32 in p-type polymer OPV power conversion efficiency, 0.91 in electron affinity, 0.76 in dielectric constant. The RoBERTa\cite{liu_roberta_2019} based transformer model uses polymer SMILES\cite{weininger_smiles_1988} as sequences of polymer repeating units and structural descriptors. The chemically aware tokenizer of the model tokenizes the SMILES as input to the TransPolymer. Bond angles, chain configuration, bond length, and electronic stability of the polymer are expected to be learned implicitly by the model.

For the prediction component of our work, we fine-tune the pretrained model on polymer datasets (table \ref{fig:downstream-datasets}) of electron affinity, bulk bandgap,  p-type polymer OPV power conversion efficiency,  electric conductivity, and dielectric constant for the polymer property regressor head, as mentioned in the TransPolymer's repository (\url{https://github.com/ChangwenXu98/TransPolymer.git}). 
Generative models can also be used to generate and search the chemical latent space in large amounts of molecular data, and hence create large synthetic databases like PolyInfo\cite{ishii_nims_2024}, PI1M\cite{doi:10.1021/acs.jcim.0c00726}, Polymer Genome\cite{kim_polymer_2018}, which are datasets or web-based user interfaces for one-time search in the synthetic latent spaces. These works have significantly accelerated the data-intensive pretraining of chemically aware models. However, these generative models do not focus on the synthetic feasibility of common reactants available in wet labs.\cite{kim_open_2023} Molecule Chef model \cite{bradshaw_model_2019} introduces de novo design (DND) for searching molecules with target property values, along with maintaining the brittle SMILES to remain valid molecules. Bradshaw et al. suggest Molecule Chef as a generative model with two components: (1) a decoder from a continuous latent space mapped to a set of easily procurable reactants (\url{https://www.emolecules.com/}) in the paper's implementation as Open Macromolecular Genome\cite{kim_open_2023}, (2) a reaction predictor model to map reactants to semantically valid molecules. Open Macromolecule Genome utilized the Molecule Chef generative model to demonstrate property targeting by augmenting log P estimations\cite{wildman_prediction_1999}, thereby providing relevant databases with constitutional repeating units (CRUs). The databases are publicly hosted at \url{https://github.com/TheJacksonLab/OpenMacromolecularGenome}. 

For the de novo design of polymers in our work, this database is extended to properties mentioned in the table \ref{fig:downstream-datasets} using the TransPolymer predictor to maintain a large synthetically accessible and semantically aligned database for finetuning the Molecule Chef's decoder. 
\begin{figure*}[!htbp] 
\centering
\includegraphics[width=0.80\textwidth]{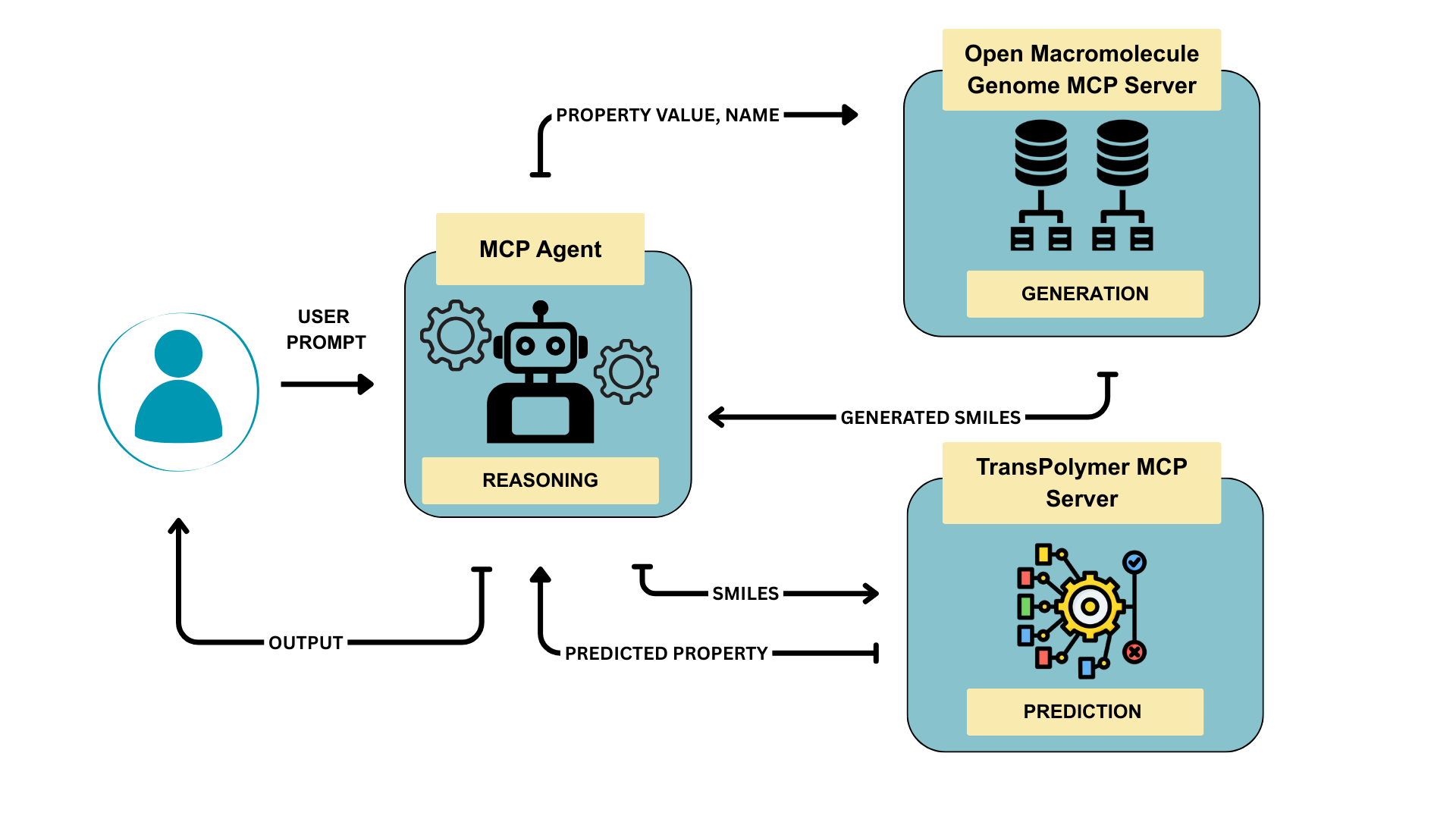} 
\caption{Workflow illustrating how each tool is sent an input and how each of the tools generates an output in the Polymer-Agent}
\label{fig:workflow}
\end{figure*}

\subsection{Model Context Protocol (MCP)}
LLMs are now capable of autonomous planning, reasoning, and resource access. These advances apply to several reasoning projects and their usage in applications towards mimicking human reasoning. \cite{xu_towards_2025} \cite{luo_mcp-universe_2025} Model Context Protocol, or MCP, also referred to as "USB-C of AI," was introduced in 2024 by Anthropic.\cite{noauthor_introducing_nodate} It integrates the recent work in LLMs and increases the adoption of LLMs in recent studies. Major AI providers like OpenAI, Google Gemini, and codebases like Cline and Cursor have used the framework to enhance their products, and hence, MCP servers have gained rapid traction.\cite{luo_mcp-universe_2025}  MCP frameworks act as an interface layer for AI sources, tools, and resources. MCP architecture works with three components: host, client, server. The host interacts with control of the tools and user requests, and is the user-facing LLM applications. The host uses the LLM to extract the user query to call the tools via the MCP client. The MCP client consists of multiple servers that can enable tools and resources assigned to them. The basic entities of MCP include: tools (defined functions), resources (databases), and prompts (templates for workflow)\cite{errico_securing_2025}\cite{noauthor_what_nodate}

\section{Agent Design}
\subsection{Task Description}

The agent serves as an end-to-end pipeline for performing property prediction and structure generation, as illustrated in figure~\ref{fig:workflow}. 


\textbf{Generation of SMILES}. For the generation of new polymer structures, it is very important to generate valid SMILES (Simplified Molecular Input Line Entry System) \cite{weininger_smiles_1988}, for polymer reactions with common reagents. The field of structure generation/ optimization is still evolving, and notable generative models, reinforcement learning, and graph optimization techniques have been implemented. \cite{qiu_-demand_2024}\cite{blaschke_reinvent_2020}\cite{bradshaw_model_2019} Kim et al developed a deep generative model for property-targeted design of synthetic polymers subject to optimization of octanol$-$water solubilities, inspired by the Bradshaw et al's generative model, which proposes a synthetic polymer design with a bag of common initial reactants.\cite{kim_open_2023}\cite{bradshaw_model_2019} The SMILES generation tool utilizes the fine-tuned model trained on the set of properties, as mentioned in table \ref{fig:downstream-datasets}. The latent space for the polymer SMILES within which the new SMILES is generated is represented in the supplementary section.

\begin{figure}[!htp] 
\centering
\includegraphics[width=0.99\textwidth]{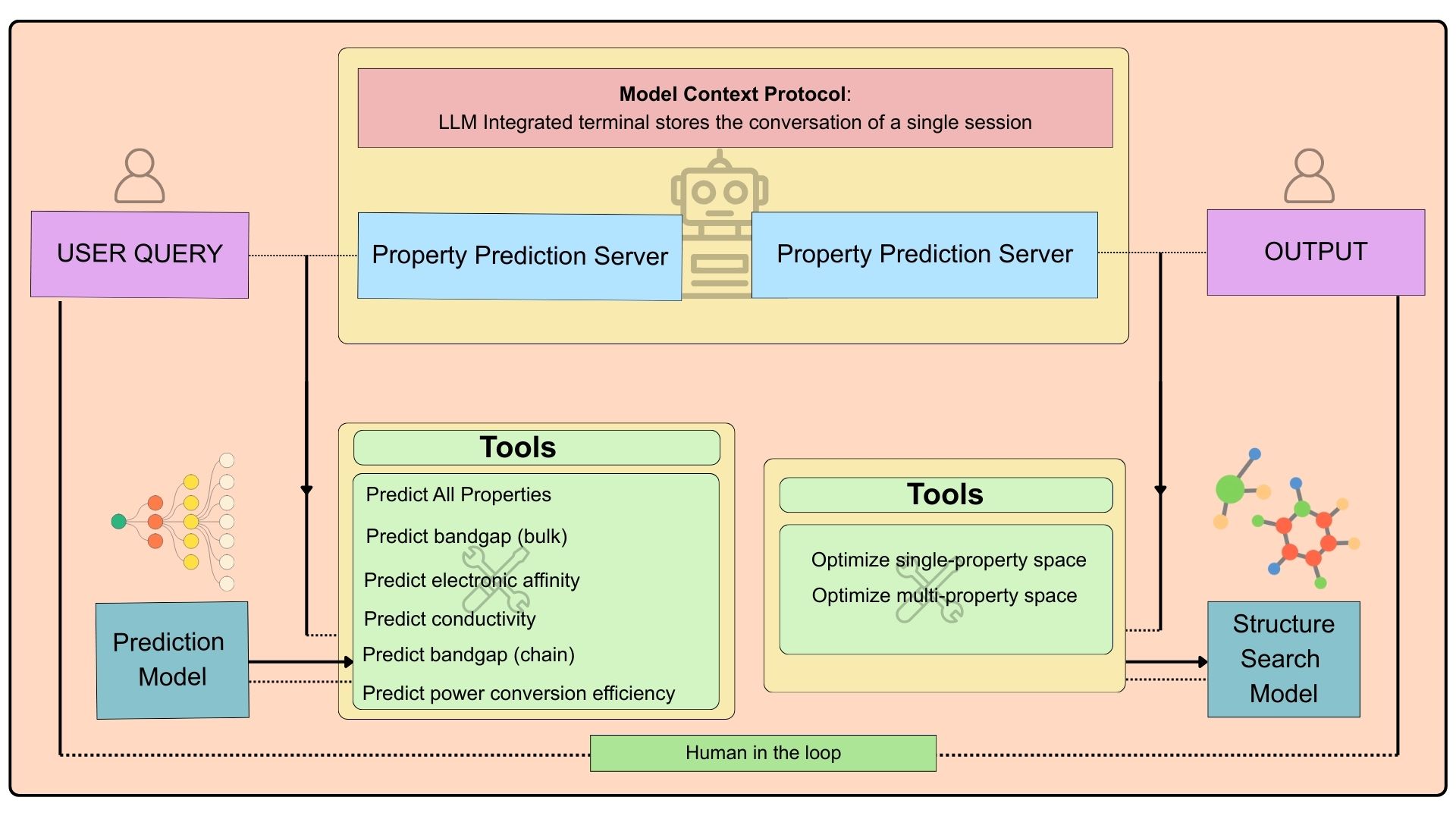} 
\caption{Tools available for usage in the MCP servers. The figure also shows how the tools access their respective utility and deliver it to the users.}
\label{fig:mcp_tools}
\end{figure}

\textbf{Property Prediction.} Machine learning algorithms, specifically graph neural networks and Transformer-based tokenization, have been used recently to predict single-task as well as multi-task properties from polymer structures as inputs, leveraging the progress achieved in the natural language domain. Multiple task-specific models, including graph neural networks and generative models (like TransPolymer, CatBERTa), enable libraries and platforms ( like JARVIS PolymerGenome, PolyId) to enable one-click search and fine-tuning of specific properties. \cite{xu_transpolymer_2023} \cite{ock_multimodal_2024}\cite{kim_polymer_2018}    \cite{wilson_polyid_2023} \cite{choudhary_jarvis_2025} The polymers are tokenized and converted to SMILES sequences where \texttt{*} is added to represent repeating units in a polymer. Other signs are also included besides the atom symbols, like \texttt{.} is used for the separation of two units of copolymers, and \textasciicircum\
 is included to indicate branches in copolymers.\cite{xu_transpolymer_2023} For each candidate molecule, Polymer-Agent uses the TransPolymer model, which has been fine-tuned for the same set of target properties as the SMILES generation task.

\begin{table}[!htbp]
    \caption{Summary of datasets for finetuning of the TransPolymer model for the downstream tasks. Table referenced from \cite{xu_transpolymer_2023}}
    \label{fig:downstream-datasets}
    \centering
    \begin{adjustbox}{width=\textwidth}
        \begin{tabular}{cccccc}
        \toprule
         Dataset & Property & \# Data & \# Augmented train data & \# Test data & Data split \\
         \midrule
         PE-\uppercase\expandafter{\romannumeral1} \cite{oyaizu_ai-assisted_2020} & conductivity & 9185 & 34803 & 146 & train-test split by year \\ 
         Egb \cite{kuenneth_polymer_2021} & bandgap (bulk) & 561 & 6443 & 113 & 5-fold cross-validation \\
         Eea\cite{kuenneth_polymer_2021} & electron affinity & 368 & 3993 & 74 & 5-fold cross-validation \\
         EPS \cite{kuenneth_polymer_2021} & dielectric constant & 382 & 4188 & 77 &5-fold cross-validation \\
         OPV \cite{nagasawa_computer-aided_2018} & power conversion efficiency & 1203 & 4810 & 241 & 5-fold cross-validation \\
        \bottomrule 
    		 
        \end{tabular}
    \end{adjustbox}
    
\end{table}
Conventional optimization techniques have used powerful algorithms to perform these tasks, but recent research shows the advantages of using autonomous constraint generation with built-in reasoning and parameter exploration \cite{zeng_llm-guided_2025}. LLM-powered agents can execute a series of optimization, prediction, and modification tasks \cite{ock_large_2025}, and in our approach, the MCP servers use the above two models to expose the generative and predictive functions as tools for iterating on polymer property prediction, structure generation, and user query handling, helping the user navigate between the two tools through human dialogs.
Specific use cases of the Polymer-Agent and sample prompts are provided in the supplementary section.
\begin{table}[!htbp]
\centering
\caption{Summary of tools exposed to the MCP Client server}
\label{tab:agent-tools}
\resizebox{\textwidth}{!}{%
\begin{tabular}{ll}
\toprule
\textbf{Task Module} & \textbf{Primary Tool Components} \\
\midrule
SMILES Generation & Molecule Chef tool, Curated Database \\
Property Prediction & TransPolymer, Molecule Chef  \\
LLM internal knowledge and reasoning & Gemini 3.0 Auto/Pro/Flash \\
Structure initial Validation & RDKit \\
\bottomrule
\end{tabular}
}
\end{table}

\subsection{Workflow}
This workflow, as shown in figure \ref{fig:mcp_tools}, works with the 3 systems concurrently - (1) LLM to maintain the reasoning of the query asked, tool exposure, and the final human-readable results. (2) The polymer SMILES generative model (3) Polymer property predictor using SMILES as inputs.

Polymer-Agent is designed to assist users in generating a plausible starting polymer structure in experimental settings or computational environments. This accelerates the polymer design experiment to predict the structure with given properties. 

The user starts with a polymer property and its value. The agent can work on a strict value as well as a range for the property value.
Polymer-Agent's MCP Client triggers the MCP server to generate structures, optimizing the latent space from two descriptors into a synthetic polymer SMILES. The SMILES obtained can be directly viewed on the Command Line Interface, and the user can request to continue or stop the generation.

After generation, TransPolymer, the prediction tool, uses SMILES to predict properties with a Transformer-based model. For using a descriptor-based database instead of the original training/finetuning corpus PI1M \cite{doi:10.1021/acs.jcim.0c00726}, the OMG database \cite{kim_open_2023} is used for finetuning the structure generation.

\begin{figure}[!hbp] 
\centering
\includegraphics[width=0.99\textwidth]{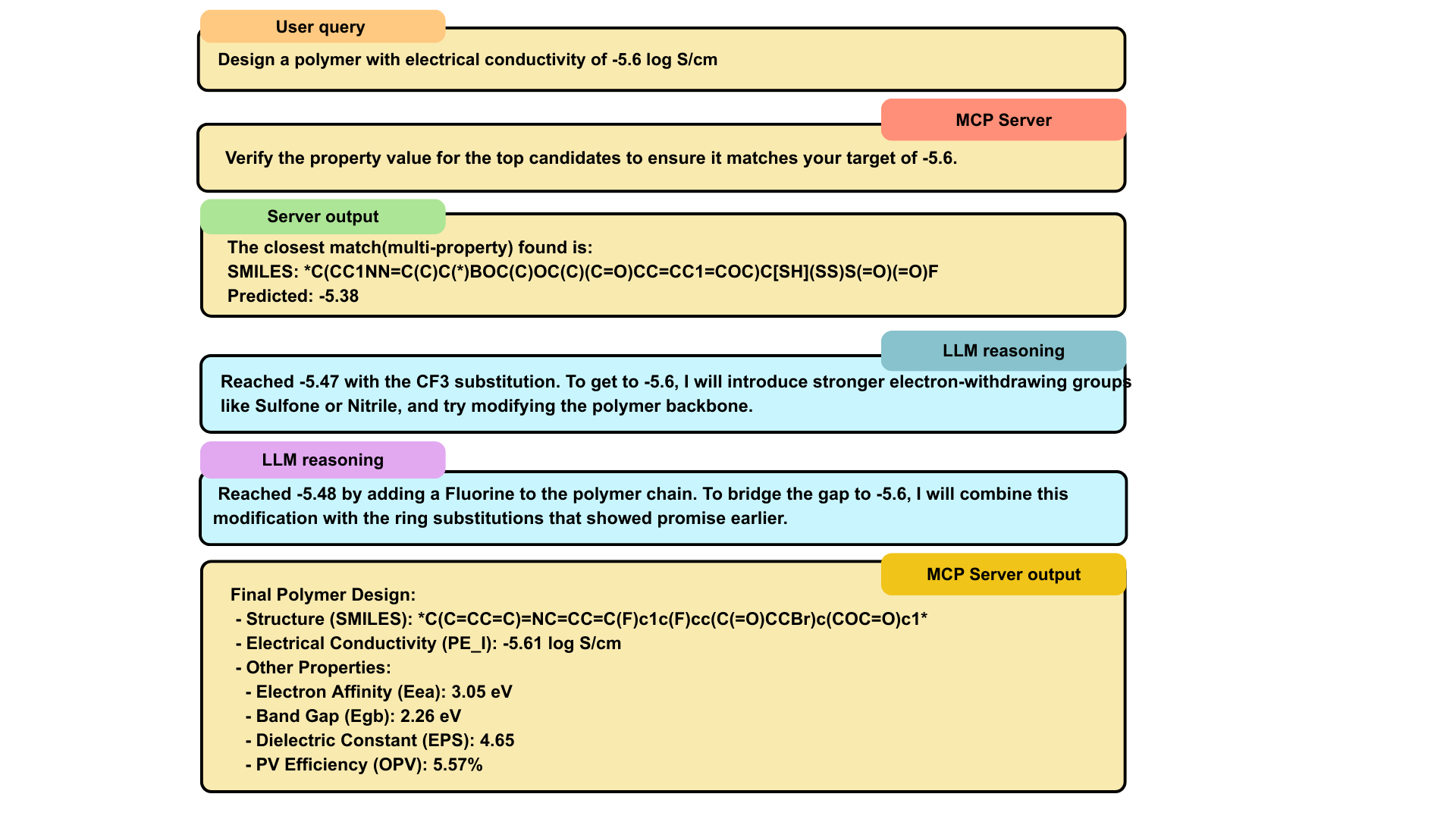} 
\caption{Workflow of the terminal responses (\textcolor{blue}{blue}: missteps taken by the LLM reasoning being crosschecked by the prediction MCP), (\textcolor{orange}{yellow}: determined steps of the MCP server) }
\label{fig:flowchart}
\end{figure}

\section{Results}
\subsection{Closed loop generation and prediction}
The AI assistant identifies the relevant tools to use for specific queries of the users. To illustrate the interface between generation and property prediction, we show the terminals' outputs (figure \ref{fig:flowchart}) in the integrated Gemini Command Line to the user's development environment.

\subsection{Generative Server}
The Molecule Chef\cite{bradshaw_model_2019} model was adapted to generate the seed molecules with property head fine-tuned on the OMG database \cite{kim_open_2023}. Building on the features of reactants and reactions from Molecule Chef, we used the synthetic database curated in this project to fine-tune the property head. The database is accessible on the GitHub page mentioned. The database consists of ~12 million distinct repeating units (CRUs) with the polymer properties of bulk bandgap, electric conductivity, p-type polymer OPV power conversion efficiency, electron affinity, and dielectric constant. SC Score\cite{coley_scscore_2018}
was chosen by Kim et al\cite{kim_open_2023} to estimate how often a reactant molecule has been
cited in published literature. The learned metric (SC score) exhibits recognition in synthetic complexity throughout a number of linear synthetic routes, measuring score from 1 to 5 with increasing complexity to fabrication (table \ref{tab:SC score}). 
\begin{figure*}[!htbp] 
\centering
\includegraphics[width=0.99\textwidth]{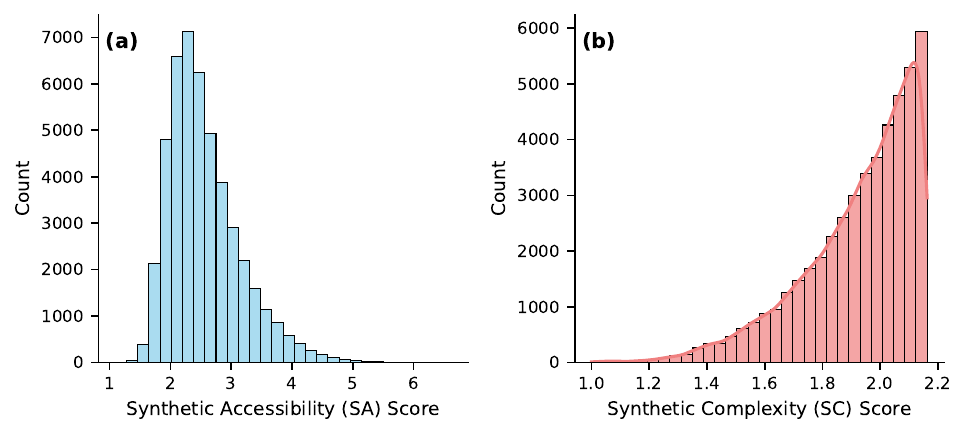} 
\caption{Representation of the Synthetic accessibility and synthetic complexity scores in OMG database.}
\label{fig:combined_score}
\end{figure*}

The Synthetic Accessibility Score (SA Score) \cite{ertl_estimation_2009} is an important metric for virtually designed molecules used in multiple databases  \cite{doi:10.1021/acs.jcim.0c00726} to numerically represent the synthesis of molecules in the laboratory. As Schuffenhaur et al. mention, molecules with a high SA score above 6, based on the distribution of SA score, are difficult to synthesize, whereas molecules with low SA score values are easily synthetically accessible. A score between 1 and 3 is perceived to be easily synthesized in reality. 
The initial OMG database, and hence the finetuning data for Polymer-Agent, display an acceptable score, ranging from 1.094 to 7.953. The histogram (see figure  \ref{fig:combined_score}) shows the distribution of the scores. 

\begin{table}[h!]
\centering
\caption{Comparison of the statistical information from the PolyInfo, PI1M and OMG datasets, to validate the latent space using SC score mentioned in the Open Macromolecule Genome article \cite{kim_open_2023}}
\begin{tabular}{lccc}
\hline
\textbf{Dataset} & \textbf{Mean SC Score}\\
\hline
OMG & 1.93 \\
PI1M  & 3.459 \\
PolyInfo &  2.16 \\
\hline
\end{tabular}
\label{tab:SC score}
\end{table}

\subsection{Property-Aware Molecular Refinement}
Key refinements to the molecular SMILES can be made using the two tools available, [Multi Property head] and [Single Property head]. The two tools differ in chemical diversity: the multi-property optimizer tends to suggest more complex structures (such as the cyano-substituted aromatic ring in the second example of table \ref{tab:Different_head} ) because it considers the shared latent space across all five properties. The multi-property optimizer allows you to add constraints (e.g., "target PE\_I of -5.3 but maintain Egb less than 4.0"), which is often more useful for actual material design. The Gemini reasoning (or any LLM integrated into the MCP server) can also account for the differences between the two methods. 
The Gemini reasoning was tested to modify particular substitutional groups in the polymer. The reasoning attempts to nudge and change the substitution groups, as shown in figures \ref{fig:flowchart}, which illustrates the chemical logic of increasing or decreasing the targeted property. The property is re-measured using the property-predictor server. However, the LLM's general reasoning modifications can lead to latent-space movements in the opposite direction, causing the target to move farther from already generated SMILES. This is flagged by the server because it is designed to always validate its prediction using the Property predictor generative model and never proceed with the LLM's reasoning as is. 
The multi-property head  optimizes the same target of electrical conductivity of -5.3 to \texttt{*CCOCCOCCCCC(Br)=CC=COCOC1CC(C\#N)CCC1OC(C=O)OC*} and the single property head optimizes the SMILES to \texttt{*}OCCOC(C)C(Br)CSO\texttt{*}, the differences in the respective properties are as illustrated in the table \ref{tab:Different_head}.

\begin{table}[h]
\caption{Comparison of the Single Property Property head. Here is the comparison for a target of electrical conductivity, -5.3}
\begin{tabularx}{\columnwidth}{lcc}
\hline
\textbf{Metrics} & \textbf{Single Property Head} & \textbf{Multi Property Head} \\
\hline
\texttt{PE\_I} (Target achieved) & -5.37 & -5.72    \\
Electron Affinity (Eea)  &  0.99 eV  & 1.34 eV \\
Chain Bandgap (Egb)  & 5.68 eV  & 4.35 eV \\
Dielectric Constant (EPS) & 3.44 & 3.59 \\
PCE (OPV)  & 4.73\%  & 4.13\% \\
\hline
\end{tabularx}
\label{tab:Different_head}
\end{table}

\subsection{Property Predictor Server}
Polymer-Agent not only provides users with targeted property search for polymers but also mitigates the issue with the black box prediction framework by relying on the generative models instead of the general reasoning of the LLM. TransPolymer \cite{xu_transpolymer_2023} employs chemically aware tokenization to embed and add descriptors such as degree of polymerization, polydispersity, and chain conformation. Unlike neural networks, Transformers depend on the self-attention mechanism, which relates tokens at different positions in a sequence to learn representations of the polymer SMILES. The comparison of the property prediction modules in the Polymer-Agent with a webpage search from the Polymer Genome: \url{http://www.polymergenome.org/} data source is shown in table \ref{tab:Prediction}.

\begin{table}[h!]
\centering
\caption{Comparison of the property prediction by an open source platform, single and multi-property head of Polymer-Agent. }
\small{\setlength{\tabcolsep}{4pt} 
\begin{tabularx}{\columnwidth}{lccc}
\hline
\textbf{Platform} & \textbf{Polymer-Agent} & \textbf{Polymer Genome \cite{kim_polymer_2018}} \\
\hline
SMILES  & *OCCOC(C)C(Br)CSO*   &  *OCCOC(C)C(Br)CSO*  \\
Electron Affinity (Eea)  &  0.99 eV & 1.0 ± 0.4 eV \\
Chain Bandgap (Egb)  & 5.68 eV  & 6.4 ± 0.5 eV\\
Dielectric Constant (EPS at 1KHz) & 3.44 &  4.54 ± 0.68 \\
Power Conversion Efficiency(OPV)  & 4.73\%  & Not available \\
\hline
\end{tabularx}}
\label{tab:Prediction}
\end{table}

Yang et al \cite{yang_novo_2024} note in their paper the presence of linear backbones, fragments of "-O-CH2-CH2". While targeting the highest conductivity of polymer in their work's data(( $5.07 \times 10^{-4}\,\mathrm{S\,cm^{-1}}$), Polymer-Agent reaches \texttt{`*N=CC=NC=C(Cl)C=C1C=CC=C(OCC)C=C1OCCOCCOC`}
( $1.07 \times 10^{-4}\,\mathrm{S\,cm^{-1}}$
) as its top candidate.
Multiple such candidates are shown in the supplementary information. Sharifi et al. \cite{sharifi_dielectric_2025}, in their prediction model of dielectric constants, show an illustrative graph for actual and predicted dielectric constants with the polymer structures. We replicate the same range of 3 - 3.5 true dielectric constants and reach similar polymers with a low SA score of 3.48, as shown in supplementary information, where Polymer-Agent reaches a Tanimoto similarity\cite{bajusz2015tanimoto} of 0.34 and a dice similarity(rdkit.Chem.AtomPairs) of 0.5 calculated on Morgan Fingerprints of \texttt{`*OCC=O`}[polymer-agent's optimised polymer] and \texttt{`NCCCCCC(=O)`} [Nylon-6].These examples illustrate the use of Polymer-Agent to generate on-demand, targeted, and novel polymers.
 
\section{Discussion}
Open polymer databases with the property features are very hard to create because of the sparse data present in the domain. It is very important that the polymer database be accessible to other researchers at no cost and with minimal effort and time. PolyInfo article with MatNavi database was published for the public in 2011\cite{otsuka_polyinfo_2011}, giving access to polymer blends, copolymers, and homopolymers extracted from the published articles on the web. Polymer Genome was introduced in the 2018 article \cite{kim_polymer_2018}, providing users with an organic computational and experimental polymer database, along with machine learning tools to predict valid SMILES properties. Joint Automated Repository for Various Integrated Simulations (JARVIS) is a unified platform for computational and machine learning models, providing users with access to databases, tools, benchmarks, etc from their published tools and research.\cite{choudhary_jarvis_2025} However, these projects are either proprietary webpages or defined structured libraries, leading to the requirement of registrations, logins, or having limited tool change access.
The Polymer-Agent generates polymer SMILES reliably from a reaction-aware generative model and open-source LLM reasoning. The current implementation executes only one such model, but users can add multiple models to compare the results as generative models continue to improve. The Polymer-Agent is a closed-loop generative model, i.e., it can generate polymer SMILES in a continuous optimization space with targeted property value specified by the user. The models can also target multiple properties in the latent space at the same time. 
An attempt at chained molecular refinements guided by LLM reasoning was made, adding or removing specific functional groups to optimise the property value towards the target. However, these modifications are not always precise and can lead to unreachable polymer SMILES; therefore, a property prediction model is used to recheck the property value before presenting the final result to the user.
In general, the most appropriate scores for polymer synthesis in real life, such as the SA Score, are used to assess the accessibility of the polymer generated by the model, as shown in the Supplementary section.

\section{Open Challenges and Future Work}
\subsection{Materials seed molecules}
Polymer databases are sparsely collected, and only a few databases have served as the sole sources for training and testing polymer informatics models. Although synthetic datasets have increased in number, there is still a need for a large generalized dataset to test and validate results from computational models.

\subsection{AI agent evaluation}
Model Context Protocol frameworks have increased the ability for multiple resources and tools to work in tandem within a single system. However, the MCP tool's calling ability has been evaluated by comparing its results with those of similar websites or applications. As MCP evaluation frameworks evolve, there is scope to integrate multiple tools and APIs for tasks such as computational characterization, running molecular dynamics simulations, and evaluating the system architecture of tool-server interactions.

\section{Conclusion}
In this study, we introduced Polymer-Agent, an LLM-powered tool orchestrating agent for automated polymer SMILES discovery. We integrated two chemically aware generative models to not only generate close-to-real polymers but also optimize the structure for user-defined property values.  The user can interact with the LLM's reasoning in natural language, and the agent can easily switch between structure-generation and property-prediction models without requiring any code changes.
The key strength of Polymer-Agent is its ability to search for target property values within a chemically valid polymer space using two defined workflows.  
This makes polymer discovery more accessible to laboratories in need of polymers with targeted properties and reduces the time required to predict new structures in wet laboratories. 

\section{Technology Use Disclosure}
We used ChatGPT and Gemini web versions to help with grammar and typographical corrections during
the preparation of this preprint manuscript and in the integration of the code. The authors have carefully reviewed, verified, and approved all content to ensure accuracy and integrity.

\section{Code Availability Statement}
The code that supports this study’s findings can be found in the following publicly available
GitHub repository: \url{https://github.com/BaratiLab/PolyAgent}

\bibliography{references}

@article{qiu_-demand_2024,
    title = {On-demand reverse design of polymers with {PolyTAO}},
    volume = {10},
    copyright = {2024 The Author(s)},
    issn = {2057-3960},
    url = {https://www.nature.com/articles/s41524-024-01466-5},
    doi = {10.1038/s41524-024-01466-5},
    language = {en},
    number = {1},
    urldate = {2025-11-30},
    journal = {npj Computational Materials},
    author = {Qiu, Haoke and Sun, Zhao-Yan},
    month = nov,
    year = {2024},
    note = {Publisher: Nature Publishing Group},
    pages = {273},
}

@article{blaschke_reinvent_2020,
    title = {{REINVENT} 2.0: {An} {AI} {Tool} for {De} {Novo} {Drug} {Design}},
    volume = {60},
    issn = {1549-9596},
    shorttitle = {{REINVENT} 2.0},
    url = {https://doi.org/10.1021/acs.jcim.0c00915},
    doi = {10.1021/acs.jcim.0c00915},
    number = {12},
    urldate = {2025-11-30},
    journal = {Journal of Chemical Information and Modeling},
    author = {Blaschke, Thomas and Arús-Pous, Josep and Chen, Hongming and Margreitter, Christian and Tyrchan, Christian and Engkvist, Ola and Papadopoulos, Kostas and Patronov, Atanas},
    month = dec,
    year = {2020},
    note = {Publisher: American Chemical Society},
    pages = {5918--5922},
}

@article{kim_open_2023,
    title = {Open {Macromolecular} {Genome}: {Generative} {Design} of {Synthetically} {Accessible} {Polymers}},
    volume = {3},
    copyright = {https://creativecommons.org/licenses/by-nc-nd/4.0/},
    issn = {2694-2453, 2694-2453},
    shorttitle = {Open {Macromolecular} {Genome}},
    url = {https://pubs.acs.org/doi/10.1021/acspolymersau.3c00003},
    doi = {10.1021/acspolymersau.3c00003},
    language = {en},
    number = {4},
    urldate = {2025-11-30},
    journal = {ACS Polymers Au},
    author = {Kim, Seonghwan and Schroeder, Charles M. and Jackson, Nicholas E.},
    month = aug,
    year = {2023},
    pages = {318--330},
}

@article{kim_polymer_2018,
    title = {Polymer {Genome}: {A} {Data}-{Powered} {Polymer} {Informatics} {Platform} for {Property} {Predictions}},
    volume = {122},
    issn = {1932-7447},
    shorttitle = {Polymer {Genome}},
    url = {https://doi.org/10.1021/acs.jpcc.8b02913},
    doi = {10.1021/acs.jpcc.8b02913},
    number = {31},
    urldate = {2025-11-30},
    journal = {The Journal of Physical Chemistry C},
    author = {Kim, Chiho and Chandrasekaran, Anand and Huan, Tran Doan and Das, Deya and Ramprasad, Rampi},
    month = aug,
    year = {2018},
    note = {Publisher: American Chemical Society},
    pages = {17575--17585},
}

@article{wilson_polyid_2023,
    title = {{PolyID}: {Artificial} {Intelligence} for {Discovering} {Performance}-{Advantaged} and {Sustainable} {Polymers}},
    volume = {56},
    issn = {0024-9297},
    shorttitle = {{PolyID}},
    url = {https://doi.org/10.1021/acs.macromol.3c00994},
    doi = {10.1021/acs.macromol.3c00994},
    number = {21},
    urldate = {2025-11-30},
    journal = {Macromolecules},
    author = {Wilson, A. Nolan and St John, Peter C. and Marin, Daniela H. and Hoyt, Caroline B. and Rognerud, Erik G. and Nimlos, Mark R. and Cywar, Robin M. and Rorrer, Nicholas A. and Shebek, Kevin M. and Broadbelt, Linda J. and Beckham, Gregg T. and Crowley, Michael F.},
    month = nov,
    year = {2023},
    note = {Publisher: American Chemical Society},
    pages = {8547--8557},
}

@article{kuenneth_polybert_2023,
    title = {{polyBERT}: a chemical language model to enable fully machine-driven ultrafast polymer informatics},
    volume = {14},
    copyright = {2023 The Author(s)},
    issn = {2041-1723},
    shorttitle = {{polyBERT}},
    url = {https://www.nature.com/articles/s41467-023-39868-6},
    doi = {10.1038/s41467-023-39868-6},
    language = {en},
    number = {1},
    urldate = {2025-12-01},
    journal = {Nature Communications},
    author = {Kuenneth, Christopher and Ramprasad, Rampi},
    month = jul,
    year = {2023},
    note = {Publisher: Nature Publishing Group},
    pages = {4099},
}

@article{liao_inverse_2025,
    title = {Inverse {Design} of {Block} {Polymer} {Materials} with {Desired} {Nanoscale} {Structure} and {Macroscale} {Properties}},
    volume = {5},
    url = {https://doi.org/10.1021/jacsau.5c00377},
    doi = {10.1021/jacsau.5c00377},
    number = {6},
    urldate = {2025-12-22},
    journal = {JACS Au},
    author = {Liao, Vinson and Jayaraman, Arthi},
    month = jun,
    year = {2025},
    note = {Publisher: American Chemical Society},
    pages = {2810--2824},
}

@article{zheng_ai-guided_2025,
    title = {{AI}-{Guided} {Inverse} {Design} and {Discovery} of {Recyclable} {Vitrimeric} {Polymers}},
    volume = {12},
    copyright = {© 2024 The Author(s). Advanced Science published by Wiley-VCH GmbH},
    issn = {2198-3844},
    url = {https://onlinelibrary.wiley.com/doi/abs/10.1002/advs.202411385},
    doi = {10.1002/advs.202411385},
    language = {en},
    number = {6},
    urldate = {2025-12-22},
    journal = {Advanced Science},
    author = {Zheng, Yiwen and Thakolkaran, Prakash and Biswal, Agni K. and Smith, Jake A. and Lu, Ziheng and Zheng, Shuxin and Nguyen, Bichlien H. and Kumar, Siddhant and Vashisth, Aniruddh},
    year = {2025},
    note = {\_eprint: https://advanced.onlinelibrary.wiley.com/doi/pdf/10.1002/advs.202411385},
    pages = {2411385},
}

@misc{sahu_encoder-decoder_2025,
    title = {An {Encoder}-{Decoder} {Foundation} {Chemical} {Language} {Model} for {Generative} {Polymer} {Design}},
    url = {http://arxiv.org/abs/2510.18860},
    doi = {10.48550/arXiv.2510.18860},
    urldate = {2025-12-22},
    publisher = {arXiv},
    author = {Sahu, Harikrishna and Xiong, Wei and Savit, Anagha and Shukla, Shivank S. and Ramprasad, Rampi},
    month = oct,
    year = {2025},
    note = {arXiv:2510.18860 [cond-mat]},
}

@article{choudhary_atomgpt_2024,
    title = {{AtomGPT}: {Atomistic} {Generative} {Pretrained} {Transformer} for {Forward} and {Inverse} {Materials} {Design}},
    volume = {15},
    shorttitle = {{AtomGPT}},
    url = {https://doi.org/10.1021/acs.jpclett.4c01126},
    doi = {10.1021/acs.jpclett.4c01126},
    number = {27},
    urldate = {2025-12-22},
    journal = {The Journal of Physical Chemistry Letters},
    author = {Choudhary, Kamal},
    month = jul,
    year = {2024},
    note = {Publisher: American Chemical Society},
    pages = {6909--6917},
}

@article{ishii_nims_2024,
    title = {{NIMS} polymer database {PoLyInfo} ({I}): an overarching view of half a million data points},
    volume = {4},
    issn = {null},
    shorttitle = {{NIMS} polymer database {PoLyInfo} ({I})},
    url = {https://doi.org/10.1080/27660400.2024.2354649},
    doi = {10.1080/27660400.2024.2354649},
    number = {1},
    urldate = {2025-12-30},
    journal = {Science and Technology of Advanced Materials: Methods},
    author = {Ishii, Masashi and Ito, Takuro and Sado, Hiroko and Kuwajima, Isao},
    month = dec,
    year = {2024},
    note = {Publisher: Taylor \& Francis
\_eprint: https://doi.org/10.1080/27660400.2024.2354649},
    pages = {2354649},
}

@article{ertl_estimation_2009,
    title = {Estimation of synthetic accessibility score of drug-like molecules based on molecular complexity and fragment contributions},
    volume = {1},
    issn = {1758-2946},
    url = {https://doi.org/10.1186/1758-2946-1-8},
    doi = {10.1186/1758-2946-1-8},
    language = {en},
    number = {1},
    urldate = {2025-12-31},
    journal = {Journal of Cheminformatics},
    author = {Ertl, Peter and Schuffenhauer, Ansgar},
    month = jun,
    year = {2009},
    pages = {8},
}

@misc{liu_roberta_2019,
    title = {{RoBERTa}: {A} {Robustly} {Optimized} {BERT} {Pretraining} {Approach}},
    shorttitle = {{RoBERTa}},
    url = {http://arxiv.org/abs/1907.11692},
    doi = {10.48550/arXiv.1907.11692},
    urldate = {2026-01-02},
    publisher = {arXiv},
    author = {Liu, Yinhan and Ott, Myle and Goyal, Naman and Du, Jingfei and Joshi, Mandar and Chen, Danqi and Levy, Omer and Lewis, Mike and Zettlemoyer, Luke and Stoyanov, Veselin},
    month = jul,
    year = {2019},
    note = {arXiv:1907.11692 [cs]},
}

@article{kuenneth_polymer_2021,
    title = {Polymer informatics with multi-task learning},
    volume = {2},
    issn = {2666-3899},
    url = {https://www.sciencedirect.com/science/article/pii/S2666389921000581},
    doi = {10.1016/j.patter.2021.100238},
    number = {4},
    urldate = {2026-01-10},
    journal = {Patterns},
    author = {Kuenneth, Christopher and Rajan, Arunkumar Chitteth and Tran, Huan and Chen, Lihua and Kim, Chiho and Ramprasad, Rampi},
    month = apr,
    year = {2021},
    pages = {100238},
}

@article{nagasawa_computer-aided_2018,
    title = {Computer-{Aided} {Screening} of {Conjugated} {Polymers} for {Organic} {Solar} {Cell}: {Classification} by {Random} {Forest}},
    volume = {9},
    shorttitle = {Computer-{Aided} {Screening} of {Conjugated} {Polymers} for {Organic} {Solar} {Cell}},
    url = {https://doi.org/10.1021/acs.jpclett.8b00635},
    doi = {10.1021/acs.jpclett.8b00635},
    number = {10},
    urldate = {2026-01-10},
    journal = {The Journal of Physical Chemistry Letters},
    author = {Nagasawa, Shinji and Al-Naamani, Eman and Saeki, Akinori},
    month = may,
    year = {2018},
    note = {Publisher: American Chemical Society},
    pages = {2639--2646},
}

@article{terzopoulou_biocompatible_2022,
    title = {Biocompatible {Synthetic} {Polymers} for {Tissue} {Engineering} {Purposes}},
    volume = {23},
    issn = {1525-7797},
    url = {https://doi.org/10.1021/acs.biomac.2c00047},
    doi = {10.1021/acs.biomac.2c00047},
    number = {5},
    urldate = {2026-01-10},
    journal = {Biomacromolecules},
    author = {Terzopoulou, Zoi and Zamboulis, Alexandra and Koumentakou, Ioanna and Michailidou, Georgia and Noordam, Michiel Jan and Bikiaris, Dimitrios N.},
    month = may,
    year = {2022},
    note = {Publisher: American Chemical Society},
    pages = {1841--1863},
}

@article{li_boosting_2025,
    title = {Boosting mechanical durability under high humidity by bioinspired multisite polymer for high-efficiency flexible perovskite solar cells},
    volume = {16},
    copyright = {2025 The Author(s)},
    issn = {2041-1723},
    url = {https://www.nature.com/articles/s41467-025-57102-3},
    doi = {10.1038/s41467-025-57102-3},
    language = {en},
    number = {1},
    urldate = {2026-01-11},
    journal = {Nature Communications},
    author = {Li, Zhihao and Jia, Chunmei and Wan, Zhi and Cao, Junchao and Shi, Jishan and Xue, Jiayi and Liu, Xirui and Wu, Hongzhuo and Xiao, Chuanxiao and Li, Can and Li, Meng and Zhang, Chao and Li, Zhen},
    month = feb,
    year = {2025},
    note = {Publisher: Nature Publishing Group},
    pages = {1771},
}

@article{toh_polymer_2021,
    title = {Polymer blends and polymer composites for cardiovascular implants},
    volume = {146},
    issn = {0014-3057},
    url = {https://www.sciencedirect.com/science/article/pii/S0014305720319662},
    doi = {10.1016/j.eurpolymj.2020.110249},
    urldate = {2026-01-11},
    journal = {European Polymer Journal},
    author = {Toh, Han Wei and Toong, Daniel Wee Yee and Ng, Jaryl Chen Koon and Ow, Valerie and Lu, Shengjie and Tan, Lay Poh and Wong, Philip En Hou and Venkatraman, Subbu and Huang, Yingying and Ang, Hui Ying},
    month = mar,
    year = {2021},
    pages = {110249},
}

@misc{noauthor_sensing_nodate,
    title = {Sensing, {Imaging}, and {Therapeutic} {Strategies} {Endowing} by {Conjugate} {Polymers} for {Precision} {Medicine} - {Shen} - 2024 - {Advanced} {Materials} - {Wiley} {Online} {Library}},
    url = {https://advanced.onlinelibrary.wiley.com/doi/10.1002/adma.202310032},
    urldate = {2026-01-11},
}

@article{li_regulating_2024,
    title = {Regulating {Li}+ transport behavior by cross-scale synergistic rectification strategy for dendrite-free and high area capacity polymeric all-solid-state lithium batteries},
    volume = {72},
    issn = {2405-8297},
    url = {https://www.sciencedirect.com/science/article/pii/S2405829724005853},
    doi = {10.1016/j.ensm.2024.103759},
    urldate = {2026-01-11},
    journal = {Energy Storage Materials},
    author = {Li, Xinyang and Feng, Jie and Li, Yanan and Li, Na and Jia, Xin and Wang, Yinshui and Ding, Shujiang},
    month = sep,
    year = {2024},
    pages = {103759},
}

@article{lutz_precision_2016,
    title = {From precision polymers to complex materials and systems},
    volume = {1},
    copyright = {2016 Macmillan Publishers Limited},
    issn = {2058-8437},
    url = {https://www.nature.com/articles/natrevmats201624},
    doi = {10.1038/natrevmats.2016.24},
    language = {en},
    number = {5},
    urldate = {2026-01-11},
    journal = {Nature Reviews Materials},
    author = {Lutz, Jean-François and Lehn, Jean-Marie and Meijer, E. W. and Matyjaszewski, Krzysztof},
    month = apr,
    year = {2016},
    note = {Publisher: Nature Publishing Group},
    pages = {16024},
}

@article{meijer_mechanical_2005,
    series = {Plenary {Lectures}},
    title = {Mechanical performance of polymer systems: {The} relation between structure and properties},
    volume = {30},
    issn = {0079-6700},
    shorttitle = {Mechanical performance of polymer systems},
    url = {https://www.sciencedirect.com/science/article/pii/S0079670005000717},
    doi = {10.1016/j.progpolymsci.2005.06.009},
    number = {8},
    urldate = {2026-01-11},
    journal = {Progress in Polymer Science},
    author = {Meijer, Han E. H. and Govaert, Leon E.},
    month = aug,
    year = {2005},
    pages = {915--938},
}

@article{nguyen_recent_2016,
    title = {Recent {Advances} in {Nanostructured} {Conducting} {Polymers}: from {Synthesis} to {Practical} {Applications}},
    volume = {8},
    copyright = {http://creativecommons.org/licenses/by/3.0/},
    issn = {2073-4360},
    shorttitle = {Recent {Advances} in {Nanostructured} {Conducting} {Polymers}},
    url = {https://www.mdpi.com/2073-4360/8/4/118},
    doi = {10.3390/polym8040118},
    language = {en},
    number = {4},
    urldate = {2026-01-11},
    journal = {Polymers},
    author = {Nguyen, Duong Nguyen and Yoon, Hyeonseok},
    month = apr,
    year = {2016},
    note = {Publisher: Multidisciplinary Digital Publishing Institute},
    pages = {118},
}

@article{zhao_multifunctional_2017,
    title = {Multifunctional {Nanostructured} {Conductive} {Polymer} {Gels}: {Synthesis}, {Properties}, and {Applications}},
    volume = {50},
    issn = {0001-4842},
    shorttitle = {Multifunctional {Nanostructured} {Conductive} {Polymer} {Gels}},
    url = {https://doi.org/10.1021/acs.accounts.7b00191},
    doi = {10.1021/acs.accounts.7b00191},
    number = {7},
    urldate = {2026-01-11},
    journal = {Accounts of Chemical Research},
    author = {Zhao, Fei and Shi, Ye and Pan, Lijia and Yu, Guihua},
    month = jul,
    year = {2017},
    note = {Publisher: American Chemical Society},
    pages = {1734--1743},
}

@article{wang_inverse_2022,
    title = {Inverse {Design} of {Materials} by {Machine} {Learning}},
    volume = {15},
    copyright = {http://creativecommons.org/licenses/by/3.0/},
    issn = {1996-1944},
    url = {https://www.mdpi.com/1996-1944/15/5/1811},
    doi = {10.3390/ma15051811},
    language = {en},
    number = {5},
    urldate = {2026-01-11},
    journal = {Materials},
    author = {Wang, Jia and Wang, Yingxue and Chen, Yanan},
    month = jan,
    year = {2022},
    note = {Publisher: Multidisciplinary Digital Publishing Institute},
    pages = {1811},
}

@article{dangayach_machine_2025,
    title = {Machine {Learning}-{Aided} {Inverse} {Design} and {Discovery} of {Novel} {Polymeric} {Materials} for {Membrane} {Separation}},
    volume = {59},
    issn = {0013-936X},
    url = {https://doi.org/10.1021/acs.est.4c08298},
    doi = {10.1021/acs.est.4c08298},
    number = {2},
    urldate = {2026-01-11},
    journal = {Environmental Science \& Technology},
    author = {Dangayach, Raghav and Jeong, Nohyeong and Demirel, Elif and Uzal, Nigmet and Fung, Victor and Chen, Yongsheng},
    month = jan,
    year = {2025},
    note = {Publisher: American Chemical Society},
    pages = {993--1012},
}

@misc{qiu_introducing_2025,
    title = {Introducing {PolySea}: {An} {LLM}-{Based} {Polymer} {Smart} {Evolution} {Agent}},
    copyright = {https://creativecommons.org/licenses/by-nc/4.0/},
    shorttitle = {Introducing {PolySea}},
    url = {https://chemrxiv.org/engage/chemrxiv/article-details/67f8d2d481d2151a02f55516},
    doi = {10.26434/chemrxiv-2025-zw65g},
    language = {en},
    urldate = {2026-01-12},
    publisher = {Chemistry},
    author = {Qiu, Haoke and Zhao, Jichun and Jing, Enzhe and Hu, Weilong and Lv, Yurun and Li, Xuefeng and Sun, Zhao-Yan},
    month = apr,
    year = {2025},
}

@article{agarwal_polymer_2025,
    title = {Polymer {Solubility} {Prediction} {Using} {Large} {Language} {Models}},
    volume = {7},
    url = {https://doi.org/10.1021/acsmaterialslett.5c00054},
    doi = {10.1021/acsmaterialslett.5c00054},
    number = {6},
    urldate = {2026-01-12},
    journal = {ACS Materials Letters},
    author = {Agarwal, Sakshi and Mahmood, Akhlak and Ramprasad, Rampi},
    month = jun,
    year = {2025},
    note = {Publisher: American Chemical Society},
    pages = {2017--2023},
}

@misc{luo_mcp-universe_2025,
    title = {{MCP}-{Universe}: {Benchmarking} {Large} {Language} {Models} with {Real}-{World} {Model} {Context} {Protocol} {Servers}},
    shorttitle = {{MCP}-{Universe}},
    url = {http://arxiv.org/abs/2508.14704},
    doi = {10.48550/arXiv.2508.14704},
    urldate = {2026-01-13},
    publisher = {arXiv},
    author = {Luo, Ziyang and Shen, Zhiqi and Yang, Wenzhuo and Zhao, Zirui and Jwalapuram, Prathyusha and Saha, Amrita and Sahoo, Doyen and Savarese, Silvio and Xiong, Caiming and Li, Junnan},
    month = aug,
    year = {2025},
    note = {arXiv:2508.14704 [cs]
version: 1},
}

@misc{noauthor_what_nodate,
    title = {What is the {Model} {Context} {Protocol} ({MCP})?},
    url = {https://modelcontextprotocol.io/docs/getting-started/intro},
    language = {en},
    urldate = {2026-01-13},
    journal = {Model Context Protocol},
}

@misc{noauthor_introducing_nodate,
    title = {Introducing the {Model} {Context} {Protocol}},
    url = {https://www.anthropic.com/news/model-context-protocol},
    language = {en},
    urldate = {2026-01-13},
}

@misc{errico_securing_2025,
    title = {Securing the {Model} {Context} {Protocol} ({MCP}): {Risks}, {Controls}, and {Governance}},
    shorttitle = {Securing the {Model} {Context} {Protocol} ({MCP})},
    url = {http://arxiv.org/abs/2511.20920},
    doi = {10.48550/arXiv.2511.20920},
    urldate = {2026-01-13},
    publisher = {arXiv},
    author = {Errico, Herman and Ngiam, Jiquan and Sojan, Shanita},
    month = nov,
    year = {2025},
    note = {arXiv:2511.20920 [cs]},
}

@misc{xu_towards_2025,
    title = {Towards {Large} {Reasoning} {Models}: {A} {Survey} of {Reinforced} {Reasoning} with {Large} {Language} {Models}},
    shorttitle = {Towards {Large} {Reasoning} {Models}},
    url = {http://arxiv.org/abs/2501.09686},
    doi = {10.48550/arXiv.2501.09686},
    urldate = {2026-01-13},
    publisher = {arXiv},
    author = {Xu, Fengli and Hao, Qianyue and Zong, Zefang and Wang, Jingwei and Zhang, Yunke and Wang, Jingyi and Lan, Xiaochong and Gong, Jiahui and Ouyang, Tianjian and Meng, Fanjin and Shao, Chenyang and Yan, Yuwei and Yang, Qinglong and Song, Yiwen and Ren, Sijian and Hu, Xinyuan and Li, Yu and Feng, Jie and Gao, Chen and Li, Yong},
    month = jan,
    year = {2025},
    note = {arXiv:2501.09686 [cs]},
}

@inproceedings{vaswani_attention_2017,
    title = {Attention is {All} you {Need}},
    volume = {30},
    url = {https://proceedings.neurips.cc/paper_files/paper/2017/hash/3f5ee243547dee91fbd053c1c4a845aa-Abstract.html},
    urldate = {2026-01-14},
    booktitle = {Advances in {Neural} {Information} {Processing} {Systems}},
    publisher = {Curran Associates, Inc.},
    author = {Vaswani, Ashish and Shazeer, Noam and Parmar, Niki and Uszkoreit, Jakob and Jones, Llion and Gomez, Aidan N and Kaiser, ukasz and Polosukhin, Illia},
    year = {2017},
}

@inproceedings{wang_smiles-bert_2019,
    address = {New York, NY, USA},
    series = {{BCB} '19},
    title = {{SMILES}-{BERT}: {Large} {Scale} {Unsupervised} {Pre}-{Training} for {Molecular} {Property} {Prediction}},
    isbn = {978-1-4503-6666-3},
    shorttitle = {{SMILES}-{BERT}},
    url = {https://dl.acm.org/doi/10.1145/3307339.3342186},
    doi = {10.1145/3307339.3342186},
    urldate = {2026-01-14},
    booktitle = {Proceedings of the 10th {ACM} {International} {Conference} on {Bioinformatics}, {Computational} {Biology} and {Health} {Informatics}},
    publisher = {Association for Computing Machinery},
    author = {Wang, Sheng and Guo, Yuzhi and Wang, Yuhong and Sun, Hongmao and Huang, Junzhou},
    month = sep,
    year = {2019},
    pages = {429--436},
}

@misc{chithrananda_chemberta_2020,
    title = {{ChemBERTa}: {Large}-{Scale} {Self}-{Supervised} {Pretraining} for {Molecular} {Property} {Prediction}},
    shorttitle = {{ChemBERTa}},
    url = {http://arxiv.org/abs/2010.09885},
    doi = {10.48550/arXiv.2010.09885},
    urldate = {2026-01-14},
    publisher = {arXiv},
    author = {Chithrananda, Seyone and Grand, Gabriel and Ramsundar, Bharath},
    month = oct,
    year = {2020},
    note = {arXiv:2010.09885 [cs]},
}

@article{schwaller_molecular_2019,
    title = {Molecular {Transformer}: {A} {Model} for {Uncertainty}-{Calibrated} {Chemical} {Reaction} {Prediction}},
    volume = {5},
    issn = {2374-7943},
    shorttitle = {Molecular {Transformer}},
    url = {https://doi.org/10.1021/acscentsci.9b00576},
    doi = {10.1021/acscentsci.9b00576},
    number = {9},
    urldate = {2026-01-14},
    journal = {ACS Central Science},
    publisher = {American Chemical Society},
    author = {Schwaller, Philippe and Laino, Teodoro and Gaudin, Théophile and Bolgar, Peter and Hunter, Christopher A. and Bekas, Costas and Lee, Alpha A.},
    month = sep,
    year = {2019},
    pages = {1572--1583},
}

@article{cao_moformer_2023,
    title = {{MOFormer}: {Self}-{Supervised} {Transformer} {Model} for {Metal}–{Organic} {Framework} {Property} {Prediction}},
    volume = {145},
    issn = {0002-7863},
    shorttitle = {{MOFormer}},
    url = {https://doi.org/10.1021/jacs.2c11420},
    doi = {10.1021/jacs.2c11420},
    number = {5},
    urldate = {2026-01-14},
    journal = {Journal of the American Chemical Society},
    publisher = {American Chemical Society},
    author = {Cao, Zhonglin and Magar, Rishikesh and Wang, Yuyang and Barati Farimani, Amir},
    month = feb,
    year = {2023},
    pages = {2958--2967},
}

@misc{agarwal_polyretro_2025,
    title = {{polyRETRO}: a {Language} {Model} {Approach} to predict {Polymerization} {Class} and {Monomer}(s) for a {Target} {Polymer}},
    shorttitle = {{polyRETRO}},
    url = {http://arxiv.org/abs/2512.05138},
    doi = {10.48550/arXiv.2512.05138},
    urldate = {2026-01-14},
    publisher = {arXiv},
    author = {Agarwal, Sakshi and Xiong, Wei and Ramprasad, Rampi},
    month = dec,
    year = {2025},
    note = {arXiv:2512.05138 [cond-mat]},
}

@article{wildman_prediction_1999,
    title = {Prediction of {Physicochemical} {Parameters} by {Atomic} {Contributions}},
    volume = {39},
    issn = {0095-2338, 1520-5142},
    url = {https://pubs.acs.org/doi/10.1021/ci990307l},
    doi = {10.1021/ci990307l},
    language = {en},
    number = {5},
    urldate = {2026-01-15},
    journal = {Journal of Chemical Information and Computer Sciences},
    author = {Wildman, Scott A. and Crippen, Gordon M.},
    month = sep,
    year = {1999},
    pages = {868--873},
}

@article{coley_scscore_2018,
    title = {{SCScore}: {Synthetic} {Complexity} {Learned} from a {Reaction} {Corpus}},
    volume = {58},
    issn = {1549-9596},
    shorttitle = {{SCScore}},
    url = {https://doi.org/10.1021/acs.jcim.7b00622},
    doi = {10.1021/acs.jcim.7b00622},
    number = {2},
    urldate = {2026-01-16},
    journal = {Journal of Chemical Information and Modeling},
    publisher = {American Chemical Society},
    author = {Coley, Connor W. and Rogers, Luke and Green, William H. and Jensen, Klavs F.},
    month = feb,
    year = {2018},
    pages = {252--261},
}

@inproceedings{otsuka_polyinfo_2011,
    title = {{PoLyInfo}: {Polymer} {Database} for {Polymeric} {Materials} {Design}},
    shorttitle = {{PoLyInfo}},
    url = {https://ieeexplore.ieee.org/document/6076416/citations},
    doi = {10.1109/EIDWT.2011.13},
    urldate = {2026-01-16},
    booktitle = {2011 {International} {Conference} on {Emerging} {Intelligent} {Data} and {Web} {Technologies}},
    author = {Otsuka, Shingo and Kuwajima, Isao and Hosoya, Junko and Xu, Yibin and Yamazaki, Masayoshi},
    month = sep,
    year = {2011},
    pages = {22--29},
}

@article{choudhary_jarvis_2025,
    title = {The {JARVIS} infrastructure is all you need for materials design},
    volume = {259},
    issn = {0927-0256},
    url = {https://www.sciencedirect.com/science/article/pii/S0927025625004069},
    doi = {10.1016/j.commatsci.2025.114063},
    urldate = {2026-01-16},
    journal = {Computational Materials Science},
    author = {Choudhary, Kamal},
    month = sep,
    year = {2025},
    pages = {114063},
}

@article{wang_molecular_2022,
    title = {Molecular contrastive learning of representations via graph neural networks},
    volume = {4},
    copyright = {2022 The Author(s), under exclusive licence to Springer Nature Limited},
    issn = {2522-5839},
    url = {https://www.nature.com/articles/s42256-022-00447-x},
    doi = {10.1038/s42256-022-00447-x},
    language = {en},
    number = {3},
    urldate = {2026-01-20},
    journal = {Nature Machine Intelligence},
    publisher = {Nature Publishing Group},
    author = {Wang, Yuyang and Wang, Jianren and Cao, Zhonglin and Barati Farimani, Amir},
    month = mar,
    year = {2022},
    pages = {279--287},
}

@article{chandrasekhar_amgpt_2024,
    title = {{AMGPT}: {A} large language model for contextual querying in additive manufacturing},
    volume = {11},
    issn = {2772-3690},
    shorttitle = {{AMGPT}},
    url = {https://www.sciencedirect.com/science/article/pii/S2772369024000409},
    doi = {10.1016/j.addlet.2024.100232},
    urldate = {2026-01-20},
    journal = {Additive Manufacturing Letters},
    author = {Chandrasekhar, Achuth and Chan, Jonathan and Ogoke, Francis and Ajenifujah, Olabode and Barati Farimani, Amir},
    month = dec,
    year = {2024},
    pages = {100232},
}

@misc{chandrasekhar_nanogpt_2025,
    title = {{NANOGPT}: {A} {Query}-{Driven} {Large} {Language} {Model} {Retrieval}-{Augmented} {Generation} {System} for {Nanotechnology} {Research}},
    shorttitle = {{NANOGPT}},
    url = {http://arxiv.org/abs/2502.20541},
    doi = {10.48550/arXiv.2502.20541},
    urldate = {2026-01-20},
    publisher = {arXiv},
    author = {Chandrasekhar, Achuth and Farimani, Omid Barati and Ajenifujah, Olabode T. and Ock, Janghoon and Farimani, Amir Barati},
    month = feb,
    year = {2025},
    note = {arXiv:2502.20541 [cs]},
}

@inproceedings{liu_mcpeval_2025,
    address = {Suzhou, China},
    title = {{MCPEval}: {Automatic} {MCP}-based {Deep} {Evaluation} for {AI} {Agent} {Models}},
    isbn = {979-8-89176-334-0},
    shorttitle = {{MCPEval}},
    url = {https://aclanthology.org/2025.emnlp-demos.27/},
    doi = {10.18653/v1/2025.emnlp-demos.27},
    urldate = {2026-01-20},
    booktitle = {Proceedings of the 2025 {Conference} on {Empirical} {Methods} in {Natural} {Language} {Processing}: {System} {Demonstrations}},
    publisher = {Association for Computational Linguistics},
    author = {Liu, Zhiwei and Qiu, Jielin and Wang, Shiyu and Zhang, Jianguo and Liu, Zuxin and Ram, Roshan and Chen, Haolin and Yao, Weiran and Heinecke, Shelby and Savarese, Silvio and Wang, Huan and Xiong, Caiming},
    editor = {Habernal, Ivan and Schulam, Peter and Tiedemann, Jörg},
    month = nov,
    year = {2025},
    pages = {373--402},
}

@misc{noauthor_taskmatrixai_nodate,
    title = {{TaskMatrix}.{AI}: {Completing} {Tasks} by {Connecting} {Foundation} {Models} with {Millions} of {APIs} {\textbar} {Intelligent} {Computing}},
    url = {https://spj.science.org/doi/10.34133/icomputing.0063},
    urldate = {2026-01-20},
}

@article{ock_multimodal_2024,
    title = {Multimodal language and graph learning of adsorption configuration in catalysis},
    volume = {6},
    copyright = {2024 The Author(s), under exclusive licence to Springer Nature Limited},
    issn = {2522-5839},
    url = {https://www.nature.com/articles/s42256-024-00930-7},
    doi = {10.1038/s42256-024-00930-7},
    language = {en},
    number = {12},
    urldate = {2026-01-20},
    journal = {Nature Machine Intelligence},
    publisher = {Nature Publishing Group},
    author = {Ock, Janghoon and Badrinarayanan, Srivathsan and Magar, Rishikesh and Antony, Akshay and Barati Farimani, Amir},
    month = dec,
    year = {2024},
    pages = {1501--1511},
}

@article{cao_machine_2024,
    title = {Machine {Learning} in {Membrane} {Design}: {From} {Property} {Prediction} to {AI}-{Guided} {Optimization}},
    volume = {24},
    issn = {1530-6984},
    shorttitle = {Machine {Learning} in {Membrane} {Design}},
    url = {https://doi.org/10.1021/acs.nanolett.3c05137},
    doi = {10.1021/acs.nanolett.3c05137},
    number = {10},
    urldate = {2026-01-20},
    journal = {Nano Letters},
    publisher = {American Chemical Society},
    author = {Cao, Zhonglin and Barati Farimani, Omid and Ock, Janghoon and Barati Farimani, Amir},
    month = mar,
    year = {2024},
    pages = {2953--2960},
}

@article{badrinarayanan_multi-peptide_2025,
    title = {Multi-{Peptide}: {Multimodality} {Leveraged} {Language}-{Graph} {Learning} of {Peptide} {Properties}},
    volume = {65},
    issn = {1549-9596},
    shorttitle = {Multi-{Peptide}},
    url = {https://doi.org/10.1021/acs.jcim.4c01443},
    doi = {10.1021/acs.jcim.4c01443},
    number = {1},
    urldate = {2026-01-20},
    journal = {Journal of Chemical Information and Modeling},
    publisher = {American Chemical Society},
    author = {Badrinarayanan, Srivathsan and Guntuboina, Chakradhar and Mollaei, Parisa and Barati Farimani, Amir},
    month = jan,
    year = {2025},
    pages = {83--91},
}

@article{badrinarayanan_mofgpt_2025,
    title = {{MOFGPT}: {Generative} {Design} of {Metal}–{Organic} {Frameworks} using {Language} {Models}},
    volume = {65},
    issn = {1549-9596},
    shorttitle = {{MOFGPT}},
    url = {https://doi.org/10.1021/acs.jcim.5c01625},
    doi = {10.1021/acs.jcim.5c01625},
    number = {17},
    urldate = {2026-01-20},
    journal = {Journal of Chemical Information and Modeling},
    publisher = {American Chemical Society},
    author = {Badrinarayanan, Srivathsan and Magar, Rishikesh and Antony, Akshay and Meda, Radheesh Sharma and Barati Farimani, Amir},
    month = sep,
    year = {2025},
    pages = {9049--9060},
}

@misc{balaji_gpt-molberta_2023,
    title = {{GPT}-{MolBERTa}: {GPT} {Molecular} {Features} {Language} {Model} for molecular property prediction},
    shorttitle = {{GPT}-{MolBERTa}},
    url = {http://arxiv.org/abs/2310.03030},
    doi = {10.48550/arXiv.2310.03030},
    urldate = {2026-01-20},
    publisher = {arXiv},
    author = {Balaji, Suryanarayanan and Magar, Rishikesh and Jadhav, Yayati and Farimani, Amir Barati},
    month = oct,
    year = {2023},
    note = {arXiv:2310.03030 [physics]},
}

@article{krenn_selfies_2022,
    title = {{SELFIES} and the future of molecular string representations},
    volume = {3},
    issn = {2666-3899},
    url = {https://www.sciencedirect.com/science/article/pii/S2666389922002069},
    doi = {10.1016/j.patter.2022.100588},
    number = {10},
    urldate = {2026-01-21},
    journal = {Patterns},
    author = {Krenn, Mario and Ai, Qianxiang and Barthel, Senja and Carson, Nessa and Frei, Angelo and Frey, Nathan C. and Friederich, Pascal and Gaudin, Théophile and Gayle, Alberto Alexander and Jablonka, Kevin Maik and Lameiro, Rafael F. and Lemm, Dominik and Lo, Alston and Moosavi, Seyed Mohamad and Nápoles-Duarte, José Manuel and Nigam, AkshatKumar and Pollice, Robert and Rajan, Kohulan and Schatzschneider, Ulrich and Schwaller, Philippe and Skreta, Marta and Smit, Berend and Strieth-Kalthoff, Felix and Sun, Chong and Tom, Gary and Falk von Rudorff, Guido and Wang, Andrew and White, Andrew D. and Young, Adamo and Yu, Rose and Aspuru-Guzik, Alán},
    month = oct,
    year = {2022},
    pages = {100588},
}

@misc{yao_react_2023,
    title = {{ReAct}: {Synergizing} {Reasoning} and {Acting} in {Language} {Models}},
    shorttitle = {{ReAct}},
    url = {http://arxiv.org/abs/2210.03629},
    doi = {10.48550/arXiv.2210.03629},
    urldate = {2026-01-21},
    publisher = {arXiv},
    author = {Yao, Shunyu and Zhao, Jeffrey and Yu, Dian and Du, Nan and Shafran, Izhak and Narasimhan, Karthik and Cao, Yuan},
    month = mar,
    year = {2023},
    note = {arXiv:2210.03629 [cs]},
}

@misc{mastouri_making_2025,
    title = {Making {REST} {APIs} {Agent}-{Ready}: {From} {OpenAPI} to {MCP} {Servers} for {Tool}-{Augmented} {LLMs}},
    shorttitle = {Making {REST} {APIs} {Agent}-{Ready}},
    url = {http://arxiv.org/abs/2507.16044},
    doi = {10.48550/arXiv.2507.16044},
    urldate = {2026-01-21},
    publisher = {arXiv},
    author = {Mastouri, Meriem and Ksontini, Emna and Kessentini, Wael},
    month = sep,
    year = {2025},
    note = {arXiv:2507.16044 [cs]},
}

@article{yang_novo_2024,
    title = {De novo design of polymer electrolytes using {GPT}-based and diffusion-based generative models},
    volume = {10},
    copyright = {2024 The Author(s)},
    issn = {2057-3960},
    url = {https://www.nature.com/articles/s41524-024-01470-9},
    doi = {10.1038/s41524-024-01470-9},
    language = {en},
    number = {1},
    urldate = {2026-01-22},
    journal = {npj Computational Materials},
    publisher = {Nature Publishing Group},
    author = {Yang, Zhenze and Ye, Weike and Lei, Xiangyun and Schweigert, Daniel and Kwon, Ha-Kyung and Khajeh, Arash},
    month = dec,
    year = {2024},
    pages = {296},
}

@article{sharifi_dielectric_2025,
    title = {Dielectric constant prediction in polymers: {A} chemical structure based approach},
    volume = {8},
    issn = {2949-8228},
    shorttitle = {Dielectric constant prediction in polymers},
    url = {https://www.sciencedirect.com/science/article/pii/S2949822825003132},
    doi = {10.1016/j.nxmate.2025.100795},
    urldate = {2026-01-22},
    journal = {Next Materials},
    author = {Sharifi, S. and Bonardd, S. and Miccio, L. A.},
    month = jul,
    year = {2025},
    pages = {100795},
}

@article{adams_human---loop_2024,
	title = {Human-in-the-loop for {Bayesian} autonomous materials phase mapping},
	volume = {7},
	issn = {2590-2393, 2590-2385},
	url = {https://www.cell.com/matter/abstract/S2590-2385(24)00006-7},
	doi = {10.1016/j.matt.2024.01.005},
	language = {English},
	number = {2},
	urldate = {2025-11-27},
	journal = {Matter},
	publisher = {Elsevier},
	author = {Adams, Felix and McDannald, Austin and Takeuchi, Ichiro and Kusne, A. Gilad},
	month = feb,
	year = {2024},
	pages = {697--709},
}

@article{xu_transpolymer_2023,
	title = {{TransPolymer}: a {Transformer}-based language model for polymer property predictions},
	volume = {9},
	copyright = {2023 The Author(s)},
	issn = {2057-3960},
	shorttitle = {{TransPolymer}},
	url = {https://www.nature.com/articles/s41524-023-01016-5},
	doi = {10.1038/s41524-023-01016-5},
	language = {en},
	number = {1},
	urldate = {2025-11-27},
	journal = {npj Computational Materials},
	publisher = {Nature Publishing Group},
	author = {Xu, Changwen and Wang, Yuyang and Barati Farimani, Amir},
	month = apr,
	year = {2023},
	pages = {64},
}

@article{weininger_smiles_1988,
	title = {{SMILES}, a chemical language and information system. 1. {Introduction} to methodology and encoding rules},
	volume = {28},
	issn = {0095-2338},
	url = {https://doi.org/10.1021/ci00057a005},
	doi = {10.1021/ci00057a005},
	number = {1},
	urldate = {2025-11-30},
	journal = {Journal of Chemical Information and Computer Sciences},
	publisher = {American Chemical Society},
	author = {Weininger, David},
	month = feb,
	year = {1988},
	pages = {31--36},
}

@article{doi:10.1021/acs.jcim.0c00726,
author = {Ma, Ruimin and Luo, Tengfei},
title = {PI1M: A Benchmark Database for Polymer Informatics},
journal = {Journal of Chemical Information and Modeling},
volume = {60},
number = {10},
pages = {4684-4690},
year = {2020},
doi = {10.1021/acs.jcim.0c00726},
    note ={PMID: 32986418},

URL = { 
    
        https://doi.org/10.1021/acs.jcim.0c00726
    
    

},
eprint = { 
    
        https://doi.org/10.1021/acs.jcim.0c00726
    
    

}

}

@misc{bradshaw_model_2019,
	title = {A {Model} to {Search} for {Synthesizable} {Molecules}},
	url = {http://arxiv.org/abs/1906.05221},
	doi = {10.48550/arXiv.1906.05221},
	urldate = {2025-11-30},
	publisher = {arXiv},
	author = {Bradshaw, John and Paige, Brooks and Kusner, Matt J. and Segler, Marwin H. S. and Hernández-Lobato, José Miguel},
	month = dec,
	year = {2019},
	note = {arXiv:1906.05221 [cs]},
}

@misc{vogel_inverse_2024,
	title = {Inverse {Design} of {Copolymers} {Including} {Stoichiometry} and {Chain} {Architecture}},
	url = {http://arxiv.org/abs/2410.02824},
	doi = {10.48550/arXiv.2410.02824},
	urldate = {2025-11-27},
	publisher = {arXiv},
	author = {Vogel, Gabriel and Weber, Jana M.},
	month = sep,
	year = {2024},
	note = {arXiv:2410.02824 [cond-mat]},
}

@article{zhao_machine-learning-assisted_2024,
	title = {Machine-learning-assisted multiscale modeling strategy for predicting mechanical properties of carbon fiber reinforced polymers},
	volume = {248},
	issn = {0266-3538},
	url = {https://www.sciencedirect.com/science/article/pii/S0266353824000253},
	doi = {10.1016/j.compscitech.2024.110455},
	urldate = {2025-11-27},
	journal = {Composites Science and Technology},
	author = {Zhao, Guomei and Xu, Tianhao and Fu, Xuemeng and Zhao, Wenlin and Wang, Liquan and Lin, Jiaping and Hu, Yaxi and Du, Lei},
	month = mar,
	year = {2024},
	pages = {110455},
}

\newpage
\section{Supplementary Work}

\begin{figure*}[!htb] 
\centering
\includegraphics[width=0.99\textwidth]{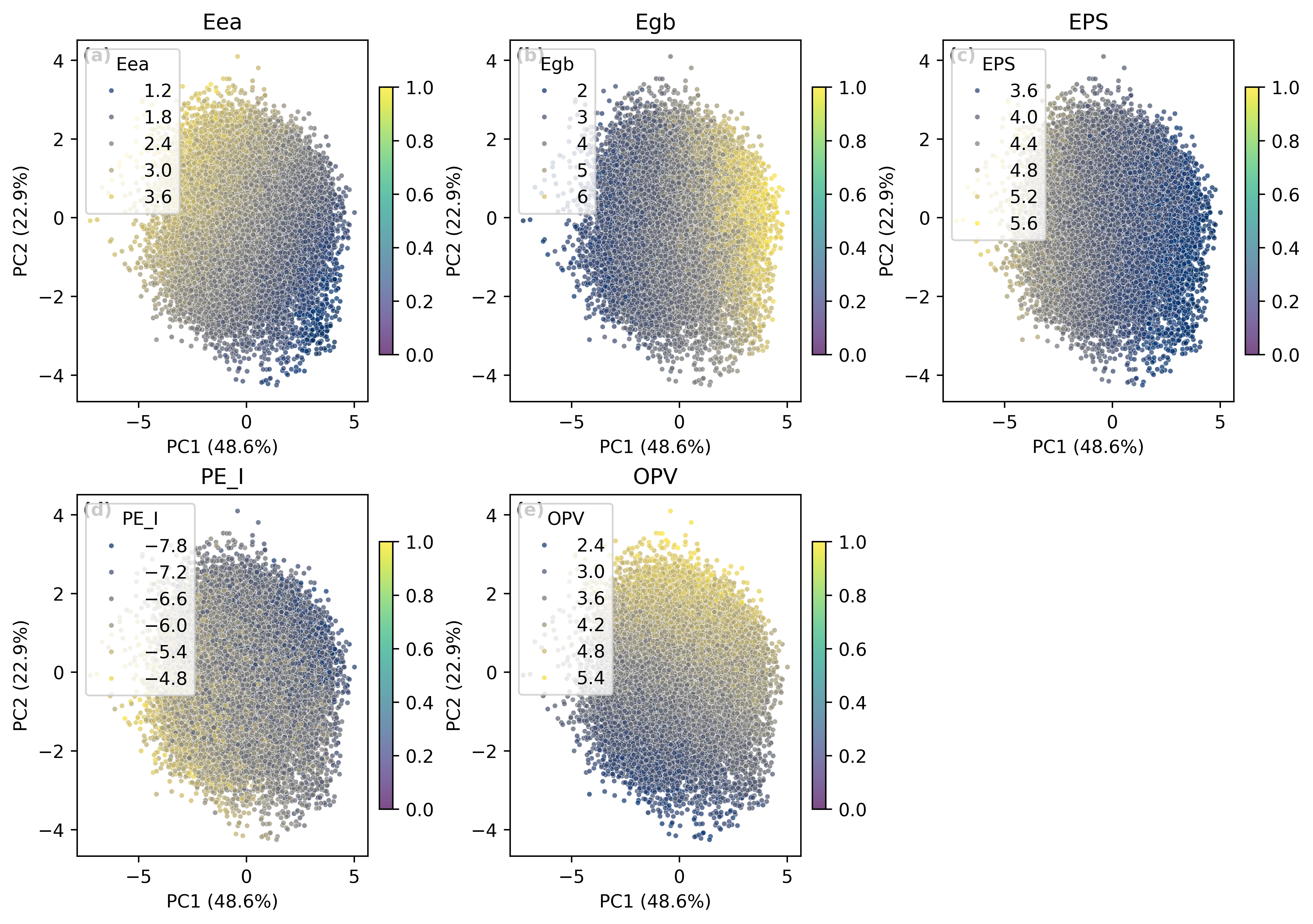} 
\caption{Representation of Principal components of the latent space of the polymer SMILES space with colour indexing of the properties.}
\label{fig:pca_properties}
\end{figure*}

Feature extraction via a variational autoencoder yields a continuous latent space that represents features of the data, in our case, polymer SMILES structures. Constitutional Repeating Units (CRUs) used for training the VAE generate the following SELFIES \cite{krenn_selfies_2022} latent spaces \ref{fig:pca_properties} with a color gradient based on the material property.\cite{kim_open_2023}
We have shown the following latent spaces, along with our database's labels, to illustrate the property gradient. 

\begin{figure}[H]
\centering
\includegraphics[width=0.99\textwidth]{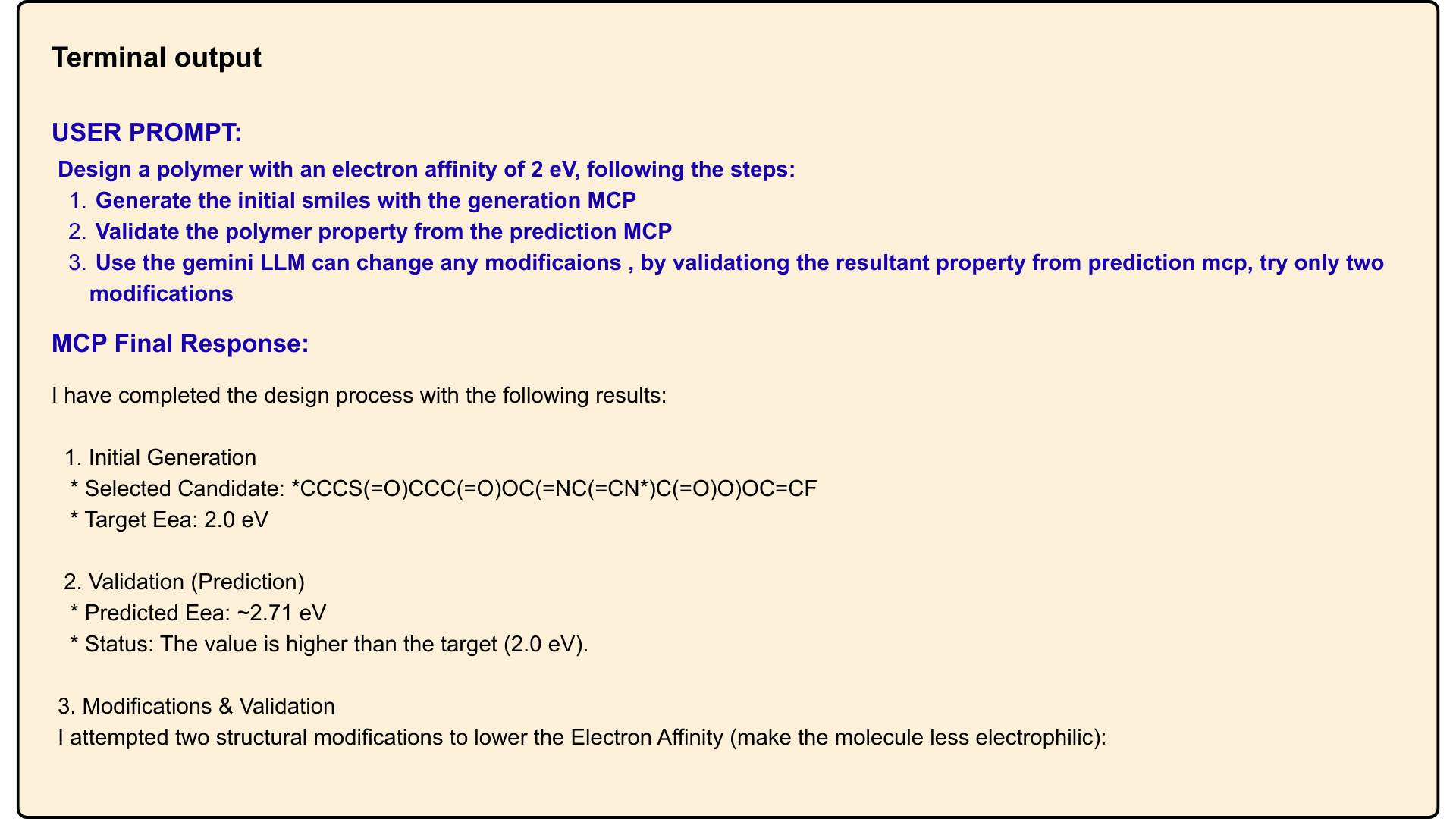}
\vspace{0.5cm}

\caption{Use-case of Polymer-Agent with final MCP response after a sample user query. Note that the user receives a prompt to move forward from the MCP server before any tool is executed.}
\label{fig:Terminal_output_1}
\end{figure}

Examples of prompts in \ref{fig:Terminal_output_1} for generating and using all the tools of the MCP servers are provided below. Users can use the same prompt template to generate valid results, as designed by the authors.

\begin{figure}[H]
\centering
\includegraphics[width=0.99\textwidth]{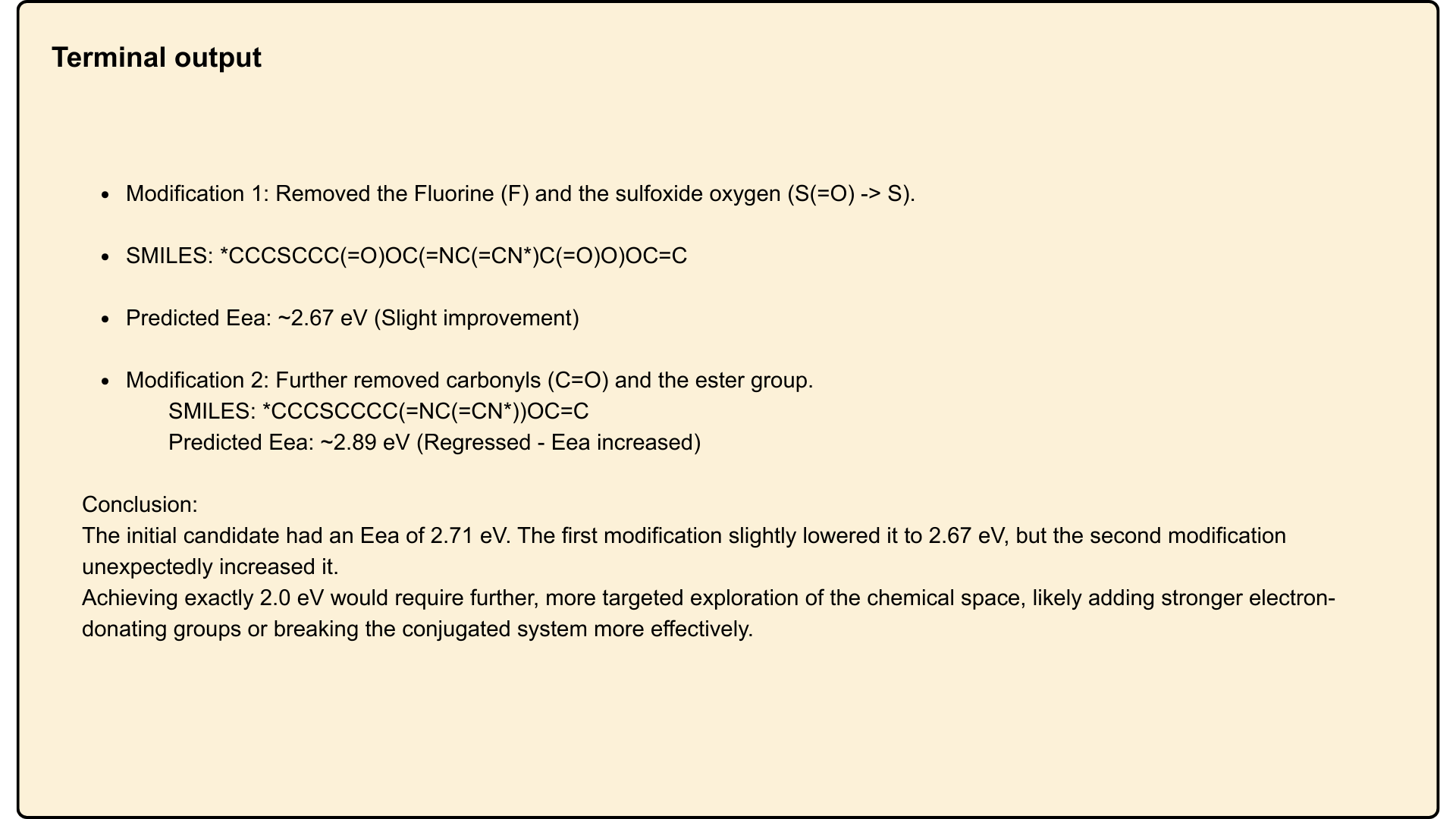}
\vspace{0.5cm}
\caption{Use-case of Polymer-Agent with final MCP response after a sample user query. Continued terminal output from figure \ref{fig:Terminal_output_1}}
\label{fig:Terminal_output_1.1}
\end{figure}

\begin{figure*}[!htbp]
\centering
\includegraphics[width=0.99\textwidth]{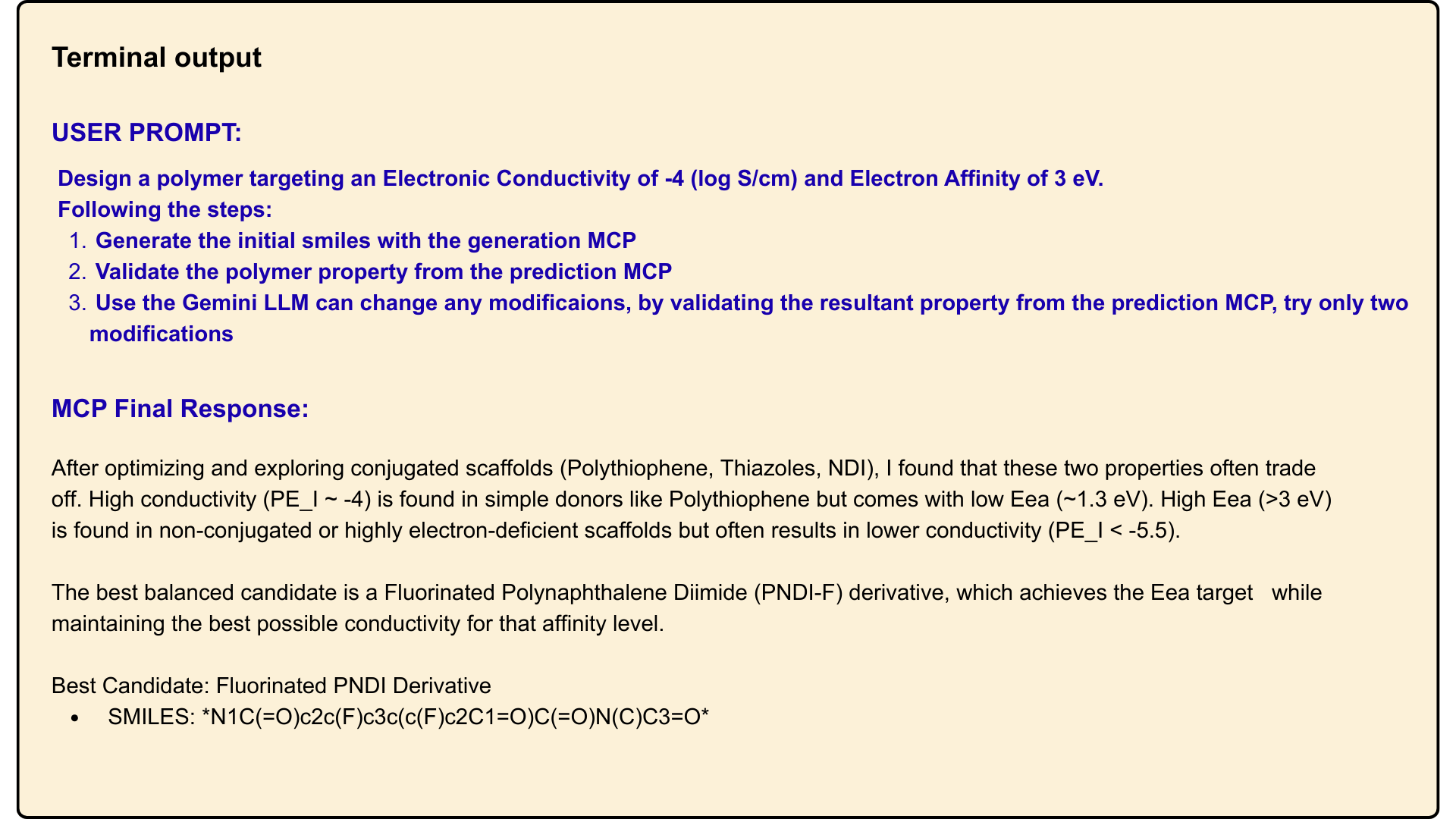}
\includegraphics[width=0.99\textwidth]{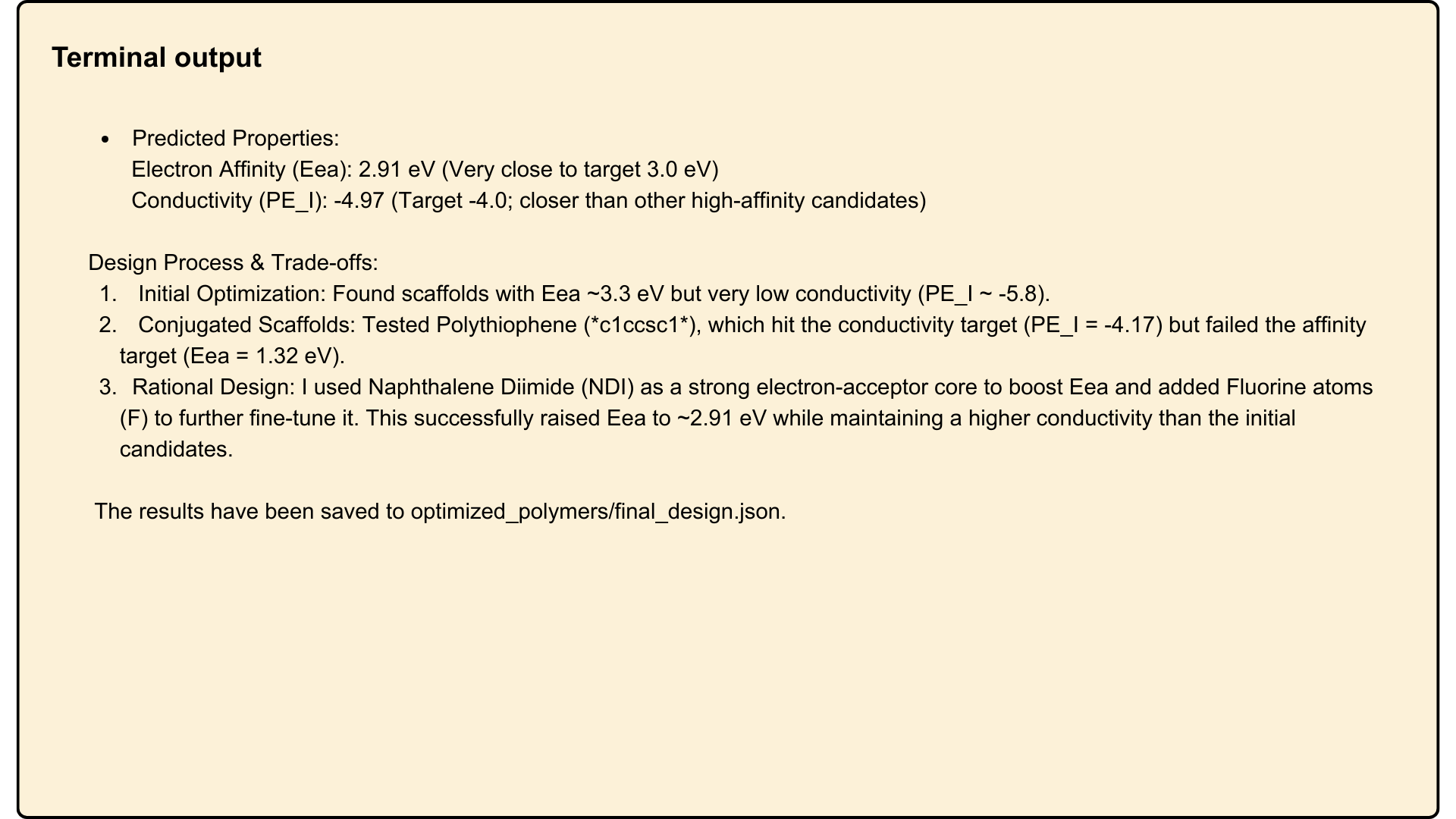}
\caption{Representation of the Terminal output for a multi-property target given to the MCP server, the results can be saved in any form as per the user in order to reuse the information obtained from the MCP server.}
\label{fig:Terminal_output_2}
\end{figure*}

Prompts in \ref{fig:Terminal_output_2} for generating and using all the tools for targeting single property optimization in Polymer-Agent.

For further validation of the agent's output, we tested the prompts using the specified tools \ref{fig:Terminal_output_3} to generate SMILES and validate the output. The same SMILES is then compared with predictions from articles using open-source models, Molecular dynamics, and other techniques in the results section.

\begin{figure}[H]
\centering
\includegraphics[width=0.99\textwidth]{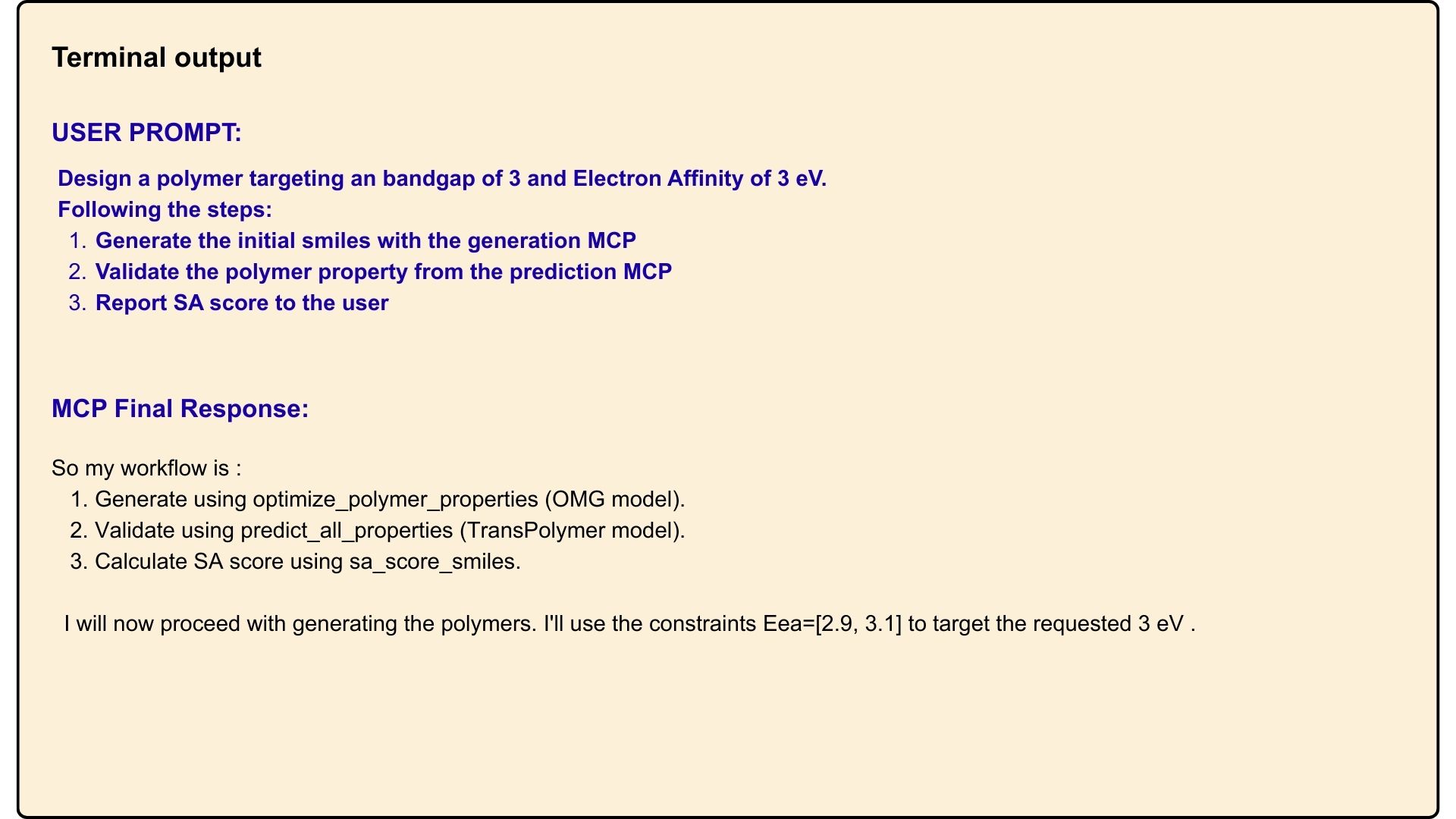}
\vspace{0.5 cm}
\includegraphics[width=0.99\textwidth]{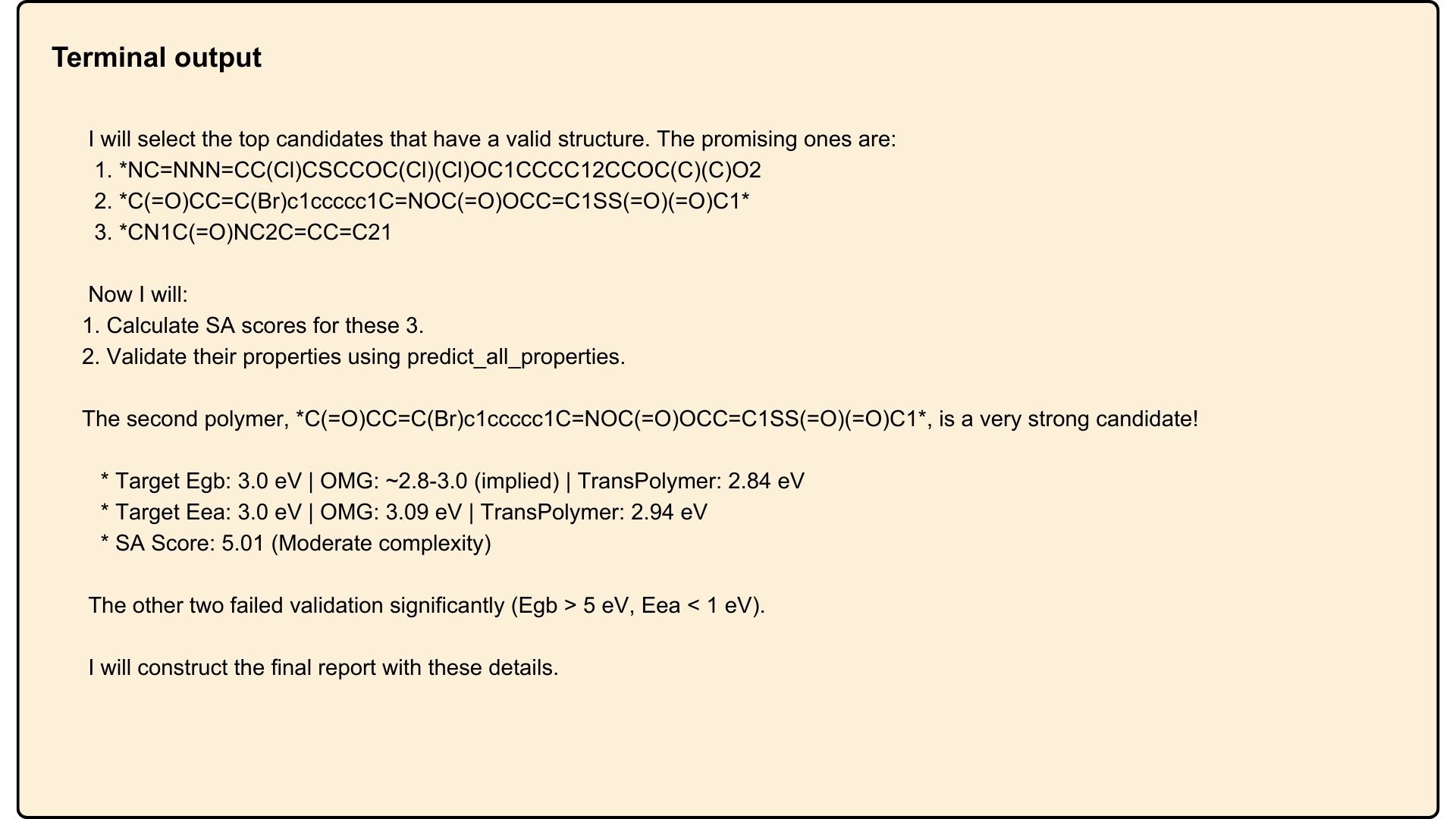}
\caption{Representation of the Terminal output for a multi-property target given to the MCP server, the results are appended with SA score for users.}
\label{fig:Terminal_output_3}
\end{figure}

\begin{figure}[H]
\centering
\includegraphics[width=0.99\textwidth]{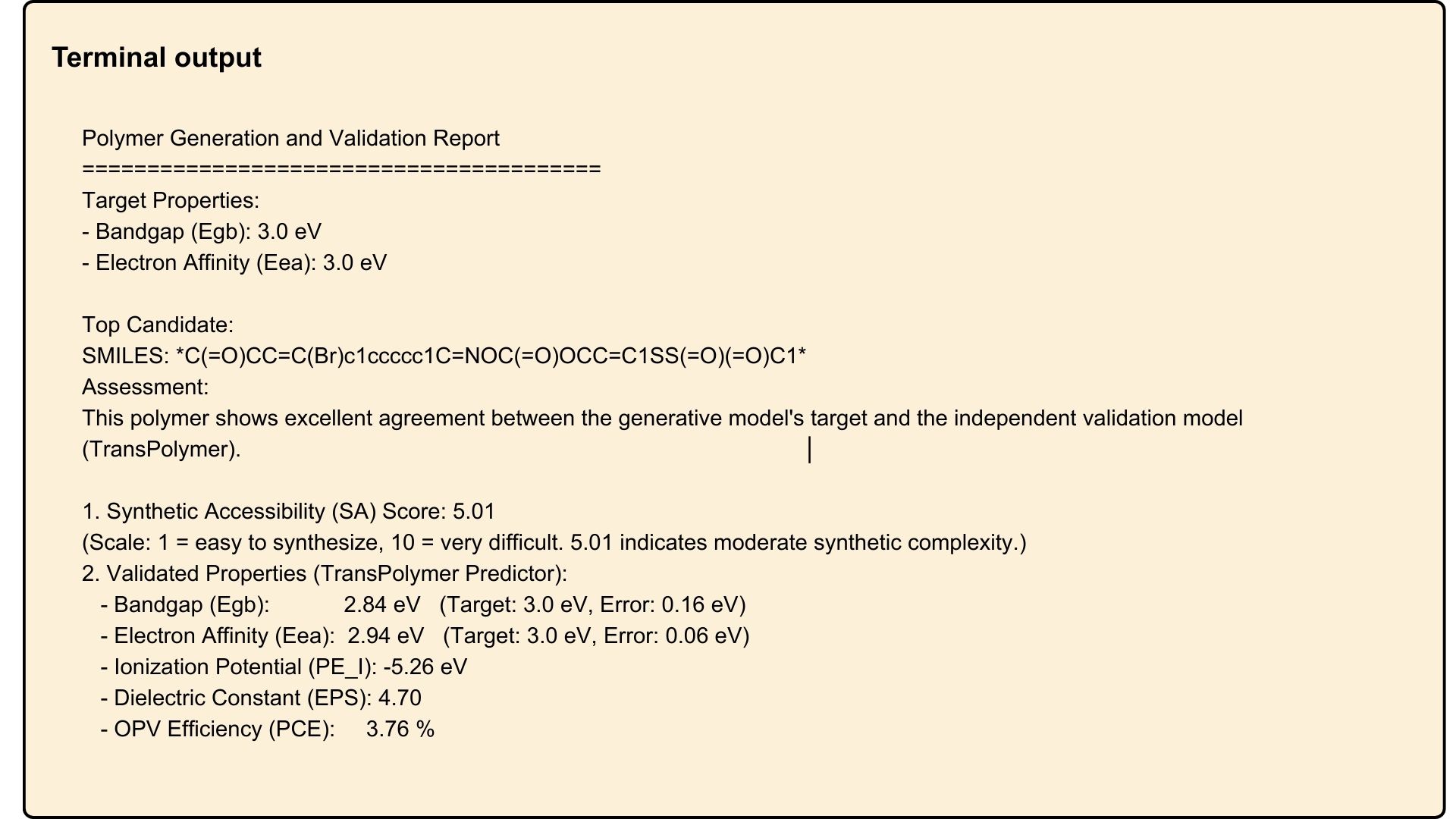}
\caption{Representation of the Terminal output for the final property report.}
\label{fig:Terminal_output_3}
\end{figure}

\begin{figure}[H]
\centering
\includegraphics[width=0.99\textwidth]{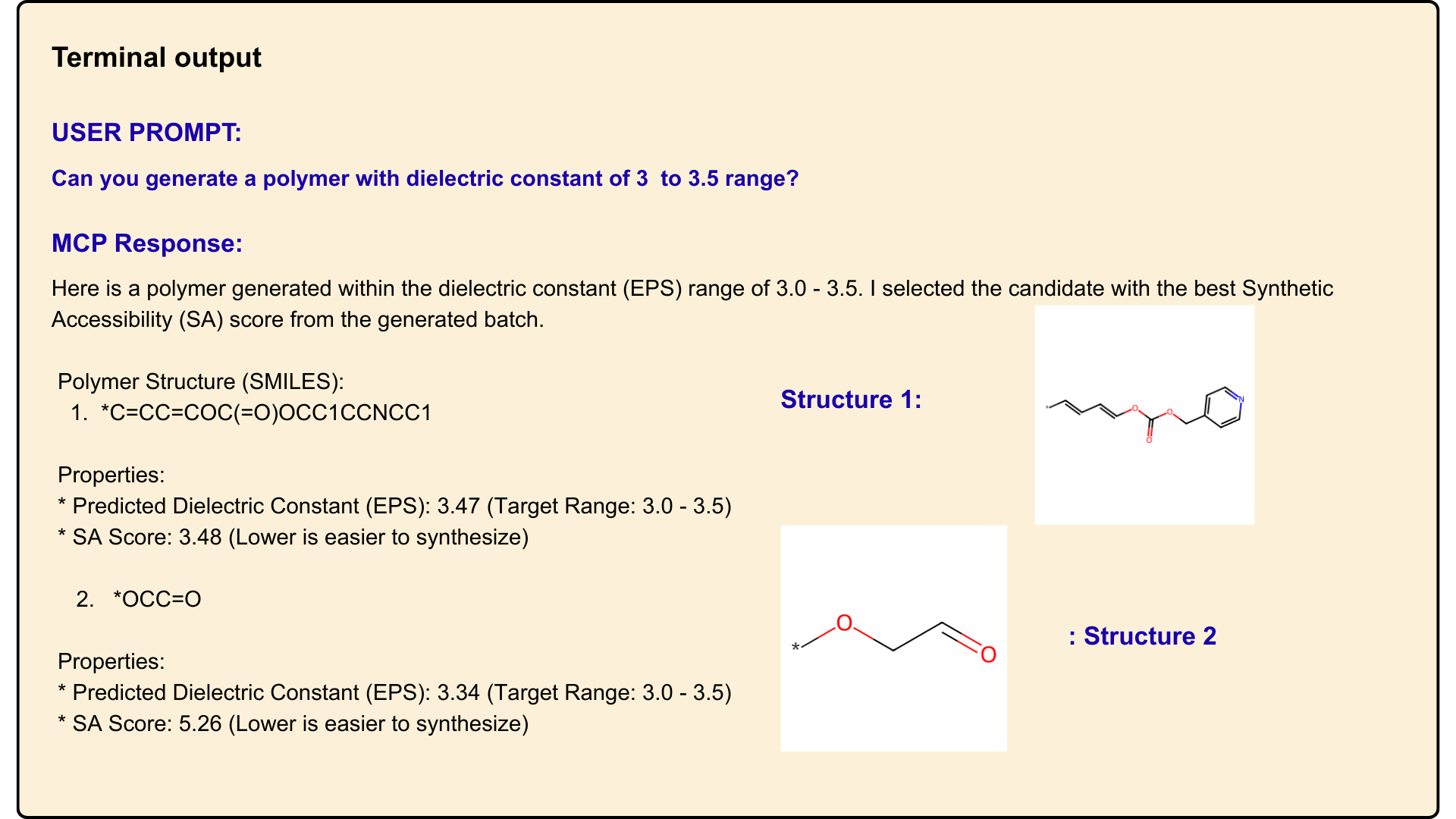}
\caption{Representation of the Terminal output for the final property report for a dielectric constant range of 3-3.5.}
\label{fig:Terminal_output_4}
\end{figure}

The results mention a user search for a dielectric constant with 3-3.5. The below terminal output\ref{fig:Terminal_output_4} shows the final one shot reply and related structures drawn with rdkit library, as image generation is not a part of Polymer-Agent yet.

\end{document}